\documentclass[10pt,twocolumn,letterpaper]{article}
\pdfoutput=1 
\DeclareUnicodeCharacter{FF0C}{U}

\usepackage{iccv}
\usepackage{times}
\usepackage{epsfig}
\usepackage{graphicx}
\usepackage{amsmath}
\usepackage{amssymb}

\usepackage{multirow, subcaption}
\usepackage{svg}
\usepackage[linesnumbered,ruled,vlined]{algorithm2e}
\SetKwInput{KwInput}{Input}
\SetKwInput{KwOutput}{Output}

\usepackage[accsupp]{axessibility}

\usepackage[american]{babel}
\usepackage{cite}
\usepackage{booktabs}
\usepackage{wasysym}
\usepackage{overpic}

\usepackage{bbding,pifont}
\usepackage{etoolbox}

\usepackage{float}
\usepackage[pagebackref=true,breaklinks=true,letterpaper=true,colorlinks,bookmarks=false]{hyperref}
\graphicspath{ {./imgs/} }
\usepackage[capitalize]{cleveref}
\crefname{section}{Sec.}{Secs.}
\Crefname{section}{Section}{Sections}
\Crefname{table}{Table}{Tables}
\crefname{table}{Tab.}{Tabs.}
\Crefname{equation}{Equation}{Equations}
\crefname{equation}{Eqn.}{Eqns.}

\usepackage[breaklinks=true,bookmarks=false]{hyperref}

\iccvfinalcopy % *** Uncomment this line for the final submission

\def\iccvPaperID{4156} % *** Enter the ICCV Paper ID here

\ificcvfinal\fi

\newtoggle{usenumbers}
\toggletrue{usenumbers}  % show circled numbers in Fig. 1 and Sec. 4.3
\newtoggle{useemoji}
\toggletrue{useemoji}  % use emoji for face
\begin{document}

%%%%%%%%% TITLE
% \title{\LaTeX\ Author Guidelines for ICCV Proceedings}
\title{{MetaF2N}: Blind Image Super-Resolution by\\Learning Efficient Model Adaptation from Faces}

\author{
Zhicun Yin$^{1}$ \quad Ming Liu$^{1(}$\Envelope$^)$ \quad Xiaoming Li$^{1}$ \quad Hui Yang$^{2}$ \quad \\
Longan Xiao$^{2}$ \quad Wangmeng Zuo$^{1,3}$ \\
$^{1}$ Harbin Institute of Technology \quad $^{2}$ Shanghai Transsion Co, Ltd \quad $^{3}$ Peng Cheng Laboratory\\
{\tt\small \href{mailto:cszcyin@outlook.com}{\color{black}cszcyin@outlook.com}, \href{mailto:csmliu@outlook.com}{\color{black}csmliu@outlook.com}, \href{mailto:csxmli@gmail.com}{\color{black}csxmli@gmail.com}, \href{mailto:wmzuo@hit.edu.cn}{\color{black}wmzuo@hit.edu.cn}}
}

\maketitle
\ificcvfinal\thispagestyle{empty}\fi

%%%%%%%%% ABSTRACT
\begin{abstract}

Due to their highly structured characteristics, faces are easier to recover than natural scenes for blind image super-resolution.
Therefore, we can extract the degradation representation of an image from the low-quality and recovered face pairs.
Using the degradation representation, realistic low-quality images can then be synthesized to fine-tune the super-resolution model for the real-world low-quality image.
However, such a procedure is time-consuming and laborious, and the gaps between recovered faces and the ground-truths further increase the optimization uncertainty.
To facilitate efficient model adaptation towards image-specific degradations, we propose a method dubbed \textbf{MetaF2N}, which leverages the contained \textbf{F}aces to fine-tune model parameters for adapting \textbf{to} the whole \textbf{N}atural image in a \textbf{Meta}-learning framework.
The degradation extraction and low-quality image synthesis steps are thus circumvented in our MetaF2N, and it requires only one fine-tuning step to get decent performance.
Considering the gaps between the recovered faces and ground-truths, we further deploy a MaskNet for adaptively predicting loss weights at different positions to reduce the impact of low-confidence areas.
To evaluate our proposed MetaF2N, we have collected a real-world low-quality dataset with one or multiple faces in each image, and our MetaF2N achieves superior performance on both synthetic and real-world datasets.
Source code, pre-trained models, and collected datasets are available at \url{https://github.com/yinzhicun/MetaF2N}.
\end{abstract}

%%%%%%%%% BODY TEXT
\section{Introduction}
\label{sec:metaf2n_introduction}

With the development of dataset construction, network design, and many other relevant methods, blind image super-resolution (SR)~\cite{Survey_InfoFuse, Survey_TRAMI22} has acquired enormous progress in recent years.
To construct pairwise low-/high-quality samples for training the image SR models, typically one can capture low- and high-quality pairs by adjusting the focal length of cameras~\cite{RealSR,DRealSR,SR-RAW} and shooting distance~\cite{City100}, or synthesize low-quality images via degradation modeling~\cite{Real-ESRGAN,BSRGAN,ReDegNet}.
However, these methods can only cover a limited and biased range of degradations, which is insufficient for real-world applications.
Besides, most existing blind image SR methods~\cite{Real-ESRGAN,BSRGAN} train a static model for all testing scenarios, greatly limiting their flexibility and generalization ability.

In order to break the restriction of limited training sets, self-supervised learning has been introduced to train a model for each low-quality image, without requiring pairwise ground-truths~\cite{DIP,ZSSR,DualSR}.
Although these methods exhibit a great deal of flexibility, they largely rely on certain priors or assumptions, showing inferior image SR performance.
Recently, Li~\etal~\cite{ReDegNet} reached a better compromise on the requirement of ground-truths and proposed ReDegNet by leveraging the faces contained in natural images.
In specific, the face regions in a real-world low-quality natural image are processed via blind face SR methods~\cite{PULSE,GPEN,GFPGAN,CodeFormer}, which have achieved appealing results thanks to the highly structured characteristics of faces.
Then, the low-quality and recovered face pairs can be utilized to model the degradations in the image.
Finally, more low-quality images are synthesized with the degradation representations for fine-tuning the SR model.
With the above design, ReDegNet~\cite{ReDegNet} is flexible to process a single image or a batch of images with diverse degradations.

Although ReDegNet~\cite{ReDegNet} achieves superior SR performance, especially in specific scenarios (\ie, fine-tuning with faces within the test image),
there are still several limitations.
On the one hand, the training procedure is time-consuming and computationally intensive, where the modules for degradation representation extraction and low-quality image synthesis are jointly optimized.
Furthermore, when the model is intended to deal with a single low-quality image, it is difficult to determine when to terminate the training process for avoiding under-fitting or over-fitting.
On the other hand, even though the faces processed by the blind face restoration methods show appealing quality, there inevitably exist gaps between the recovered faces and the real ground-truths.
Such gaps may affect the degradation representation accuracy and further hamper the training of the SR model.
These problems have a large impact on the SR performance, especially when the model is fine-tuned for image-specific super-resolution.

To remedy the aforementioned problems regarding training efforts and data gaps, we propose an efficient and effective method dubbed MetaF2N.
In specific, the SR model directly fine-tunes the parameters from the low-quality and recovered face pairs, then uses the fine-tuned parameters to process the whole natural image.
As such, the cumbersome degradation extraction and low-quality image synthesis steps are circumvented in our method, which avoids the interference of degradation modeling errors.
In order to stabilize and accelerate the fine-tuning process during inference, we adopt the model-agnostic meta-learning~\cite{MAML} framework in the training phase, resulting in a much more practical model adaptation procedure that can be finished in a single fine-tuning step.

Additionally, considering the gaps between low-quality faces and that processed by blind face restoration methods, we argue that treating all pixels equally will magnify the effect of the errors.
Therefore, we further deploy a MaskNet, which adaptively predicts loss weights for different positions, to fulfill that the recovered face regions more similar to the ground-truths are assigned with higher loss weights.
Ideally, the weight map should be extracted from the recovered faces and the ground-truths.
Unfortunately, the ground-truths are unavailable during inference.
{Considering that predicting the loss weight map is similar to the image quality assessment (IQA) task, following the idea of degraded-reference IQA} methods~\cite{DRIQA_ICCV21,DRIQA_TIP23}, the MaskNet instead takes the low-quality and recovered faces as input.
According to our observation, typically the areas closer to the ground-truth are assigned with larger weights and vice versa, which is consistent with the intuitions.

For evaluating the proposed MetaF2N, we have constructed several synthetic datasets based on {FFHQ}~\cite{FFHQ} and {CelebA}~\cite{CelebA}.
To further show the effectiveness under real-world scenarios, we have also collected a real-world low-quality image dataset from the Internet and existing datasets, namely RealFaces200 (RF200), which contains a single face or multiple faces in each image.
Extensive experiments show that our MetaF2N achieves superior performance on both synthetic and real-world datasets.

In summary, the contribution of this paper includes,
\begin{itemize}
    \item We propose an efficient and effective method dubbed MetaF2N for blind image super-resolution, which takes advances of blind face restoration models and learns model adaptation from the face regions with only one fine-tuning step.
 
    \item Considering that the gaps between faces recovered by blind face restoration methods and the ground-truth ones may result in an inaccurate degradation modeling, a MaskNet is introduced for confidence prediction to mitigate the effect of the gaps.
    \item A real-world low-quality image dataset with one or multiple faces in each image is collected, which will be helpful for face-guided blind image super-resolution.
\end{itemize}

%-------------------------------------------------------------------------
\section{Related Work}
\label{sec:metaf2n_related_work}

Recent efforts on blind image SR are mainly devoted to network design and data construction.
We recommend \cite{Survey_InfoFuse,Survey_TRAMI22} for a comprehensive review of blind image SR, and focus on the most relevant methods of data construction, which are orthogonal to network design.
We also briefly review recent advances in blind face restoration and meta-learning, which settle the foundation of this work.

\subsection{Blind Image Super-Resolution}

\noindent\textbf{Pairwise Data.}
{Recent methods have made great efforts towards pairwise datasets that cover a wider range of real-world degradations and are better aligned.
The most intuitive way is to collect pairwise samples.
For example,}
Chen~\etal~\cite{City100} collected a {City100} dataset by a DSLR and a smartphone via focal length and shooting distance adjustment, respectively.
Since {City100} was captured via postcards ignoring the geometries, Cai~\etal~\cite{RealSR} and Wei~\etal~\cite{DRealSR} proposed to capture image pairs in real-world scenes and collected two datasets {RealSR} and {DRealSR}.
Joze~\etal~\cite{ImagePairs} instead captured a dataset {ImagePairs} by embedding two cameras with different resolutions into a beam splitter.
Another approach is to synthesize low-quality images from high-quality ones.
{Early methods~\cite{IKC,DAN,ESRGAN,RealSR-Kernel} utilized classical degradation models composed of blur, down-sampling, noise, \etc, which are {insufficient} to model the real-world degradations.}
Zhang~\etal~\cite{BSRGAN} and Wang~\etal~\cite{Real-ESRGAN} proposed to mimic real-world degradations via random combinations of different degradations.
Albeit their progress on general image SR, these methods still cover a limited and biased range of real-world degradations.

\vspace{0.5em}
\noindent\textbf{Unpaired Data.}
For better utilizing unpaired low-quality images that are easier to collect, some methods~\cite{CinCGAN,FSSR,DASR} proposed to extract the degradation representations or directly learn low-quality image synthesis in an unsupervised manner.
Some methods~\cite{ZSSR,DIP} utilized the texture recurrence across different scales or the priors embraced in the model learning process, and achieved image-specific blind image super-resolution in a self-supervised scheme.
Purely relying on the unpaired samples or solely the low-quality ones, these methods may cover more degradations yet have unsatisfactory performance in most cases.
As a compromise, Li~\etal~\cite{ReDegNet} utilized the highly structured faces in real-world low-quality images, and extracted degradation representations from the low-quality faces and their counterparts recovered by blind face restoration methods.
In this work, we take a step forward to solve the problems of Li~\etal~\cite{ReDegNet} and propose a MetaF2N framework for effective blind image SR of LR natural images containing faces.

\subsection{Blind Face Restoration}
The diverse structures and complicated degradation jointly exacerbate the difficulties of blind natural image SR.
In contrast, recent methods tend to leverage the structure information of specific images (\eg, faces~\cite{GFRNet,ASFFNet,DFDNet,DMDNet,CodeFormer,GPEN} and text~\cite{MARCONet}), which have exhibited superior performance.
Li~\etal~\cite{GFRNet,ASFFNet} suggested using high-quality image(s) of the same person to provide personalized guidance.
More general face restoration methods focused on constructing dictionaries~\cite{DFDNet,DMDNet,gu2022vqfr,wang2022restoreformer}, semantic maps~\cite{chen2021progressive}, or codebooks~\cite{CodeFormer}.
Recently, the generative structure prior based methods~\cite{GFPGAN,GPEN,GLEAN,zhu2022blind} leveraged the representative pre-trained StyleGAN models~\cite{StyleGAN2} and showed tremendous improvement in restoring fine-grained textures.
{Compared with natural images, these methods showed great generalization abilities, even in the presence of unknown degradations.}
This property has \if 0 inspired \fi {motivated} us to employ face images on the inner loop of meta-learning to benefit {the optimization toward better natural image SR.}

\subsection{Image Super-Resolution with Meta-Learning}

Meta-learning aims to learn adaptation to new domains or tasks by leveraging existing ones, which has demonstrated excellent generalization ability in many tasks~\cite{vinyals2016matching,sun2019meta,sung2018learning,oreshkin2018tadam,mishra2017simple,grant2018recasting,MAML}.
Early works like MZSR~\cite{MZSR} and MLSR~\cite{MLSR} utilized meta-learning techniques to obtain a better initialization of model parameters that can be efficiently adapted to new low-quality images.
Meta-KernelGAN~\cite{Meta-KernelGAN} utilized meta-learning for efficient kernel estimation, which can be combined with non-blind SR methods.
As one can see, MZSR~\cite{MZSR} and MLSR~\cite{MLSR} heavily rely on known degradations to construct inner loop supervisions, while Meta-KernelGAN is {also restricted due to that the degradation representations formulated by kernels cannot well describe many real-world degradations}.
{In this work, we adopt face regions as inner loop supervision and get rid of explicit degradation representations, which is more practical for real-world scenarios.}

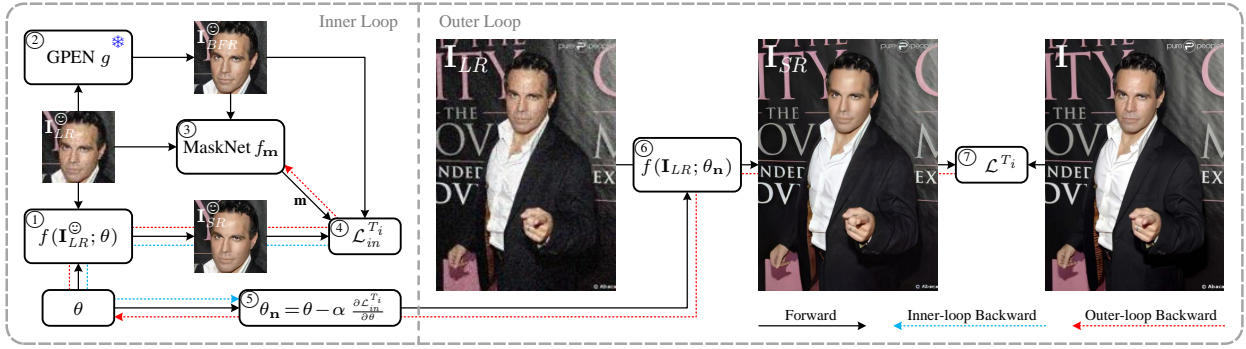
\begin{figure*}
    \centering
    \scriptsize
    \setlength\abovecaptionskip{0.5\baselineskip} 
    \setlength\belowcaptionskip{0.5\baselineskip}
    \if 0
    \begin{overpic}[grid=false,percent,width=.95\linewidth]{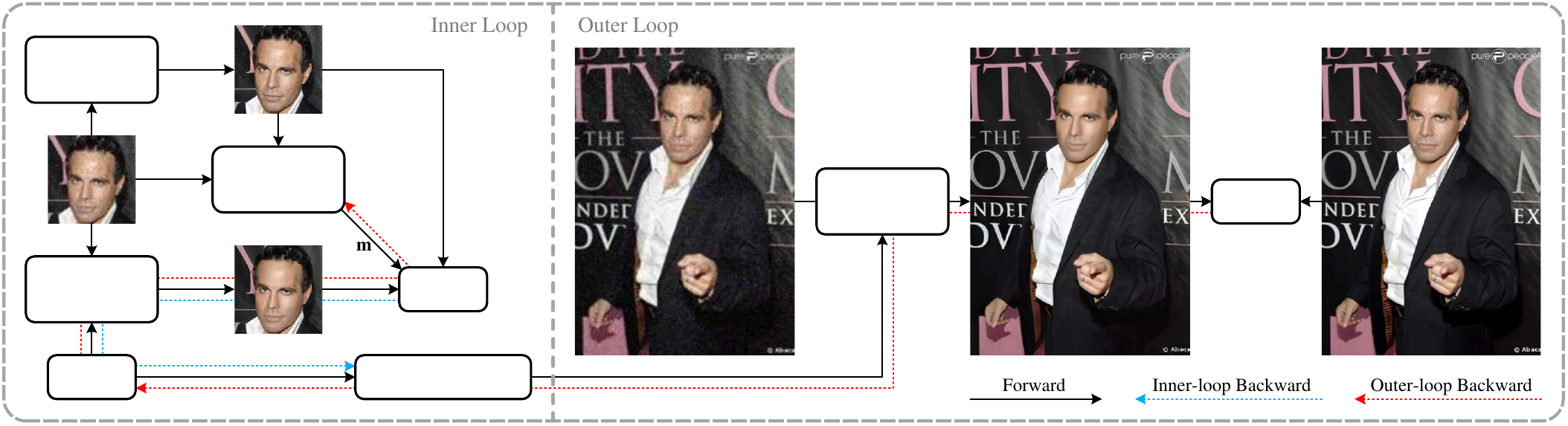}
        \put(3.5 , 22  ){GPEN $\mathit{g}$}
        \put(13.9, 16  ){MaskNet $\mathit{f}_\mathbf{m}$}
        \put(15.2, 14.5){\cref{eqn:metaf2n_masknet}}
        %\put(2.4 , 8.2 ){SR Model $\mathit{f}$}
        \put(2.8 , 8.2 ){$\mathit{f}(\mathbf{I}^{\iftoggle{useemoji}{\scriptsize\smiley{}}{\mathit{face}}}_\mathit{LR};\theta)$}
        \put(27  , 8.2 ){$\mathcal{L}_\mathit{in}$}
        \put(5.5 , 2.6 ){$\theta$}
        \put(23.1 , 2.6 ){$\theta_\mathbf{n}\!=\!\theta\!-\!\alpha\tfrac{\partial\mathcal{L}_\mathit{in}}{\partial\theta}$}
        % \put(52.7, 13.8){SR Model $\mathit{f}$}
        \put(52.7, 13.8){$\mathit{f}(\mathbf{I}_\mathit{LR};\theta_\mathbf{n})$}
        \put(79.3, 13.8){$\mathcal{L}$}
    \end{overpic}
    \fi
    \begin{overpic}[grid=false,percent,width=.95\linewidth]{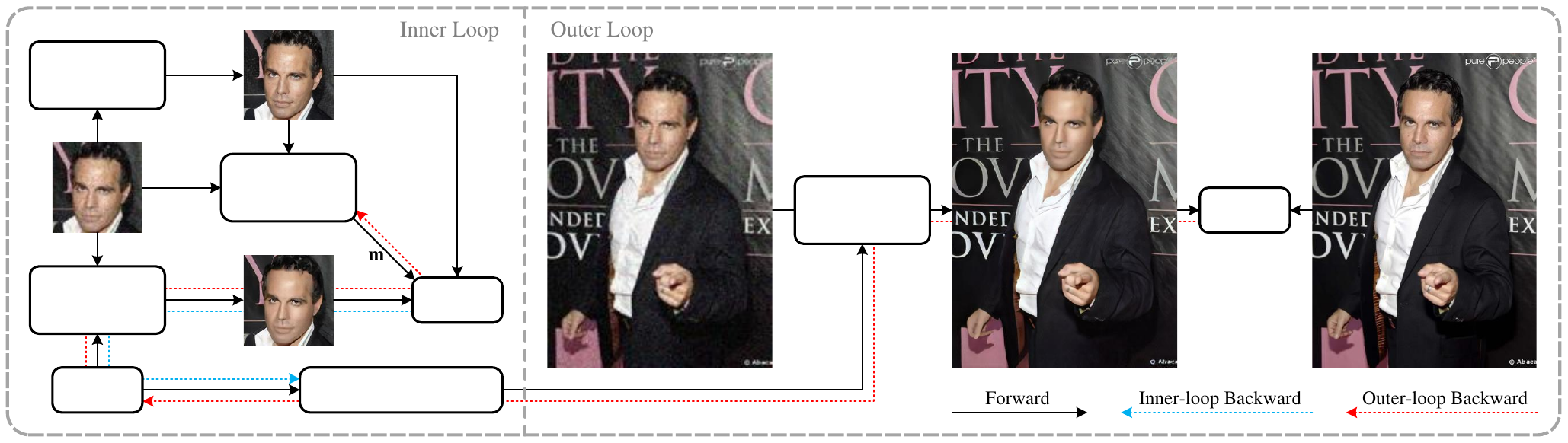}
        \put(8.9 , 24.2){\color{blue}\ding{100}}%\SnowflakeChevron}
        \put(3.7 , 23  ){GPEN $\mathit{g}$}
        %\put(14.6, 16.5){MaskNet $\mathit{f}_\mathbf{m}$}
        %\put(15.9, 14.8){\cref{eqn:metaf2n_masknet}}
        \put(14.6, 15.8){MaskNet $\mathit{f}_\mathbf{m}$}
        %\put(2.4 , 8.2 ){SR Model $\mathit{f}$}
        \put(3.1 , 8.7 ){$\mathit{f}(\mathbf{I}^{\iftoggle{useemoji}{\scriptsize\smiley{}}{\mathit{face}}}_\mathit{LR};\theta)$}
        \put(28.1, 8.7 ){$\mathcal{L}^\mathit{T_{i}}_\mathit{in}$}
        \put(5.8 , 2.8 ){$\theta$}
        \put(20.7, 2.8 ){$\theta_\mathbf{n}\!=\!\theta\!-\!\alpha$ \scalebox{0.7}{$\tfrac{\partial\mathcal{L}^\mathit{T_{i}}_\mathit{in}}{\partial\theta}$}}
        % \put(52.7, 13.8){SR Mode\scalebox{0.8}{l $\mathit{f}$}
        \put(51.4, 14.3){$\mathit{f}(\mathbf{I}_\mathit{LR};\theta_\mathbf{n})$}
        \put(78.8, 14.3){$\mathcal{L}^\mathit{T_{i}}$}

        \put(3.5 , 17.4){\color{white}$\mathbf{I}^{\iftoggle{useemoji}{\scriptsize\smiley{}}{\mathit{face}}}_\mathit{LR}$}
        \put(15.7, 24.6){\color{white}$\mathbf{I}^{\iftoggle{useemoji}{\scriptsize\smiley{}}{\mathit{face}}}_\mathit{BFR}$}
        \put(15.7, 10.3){\color{white}$\mathbf{I}^{\iftoggle{useemoji}{\scriptsize\smiley{}}{\mathit{face}}}_\mathit{SR}$}
        \put(35.5, 22.7){\large\color{white}$\mathbf{I}_\mathit{LR}$}
        \put(61.2, 22.7){\large\color{white}$\mathbf{I}_\mathit{SR}$}
        \put(84.5, 22.7){\large\color{white}$\mathbf{I}$}
%        \put(45.1, 6.7){\large\color{white}$\mathbf{I}_\mathit{LR}$}
%        \put(71.1, 6.7){\large\color{white}$\mathbf{I}_\mathit{SR}$}
%        \put(95.1, 6.7){\large\color{white}$\mathbf{I}$}

        \iftoggle{usenumbers}{
            \put(1.9 , 10.1){\scriptsize\textcircled{\tiny 1}}
            \put(1.9 , 24.5){\scriptsize\textcircled{\tiny 2}}
            \put(14.1, 17.3){\scriptsize\textcircled{\tiny 3}}
            \put(26.3, 9.4 ){\scriptsize\textcircled{\tiny 4}}
            \put(19.1, 3.7 ){\scriptsize\textcircled{\tiny 5}}
            \put(50.7, 15.8){\scriptsize\textcircled{\tiny 6}}
            \put(76.6, 15.1){\scriptsize\textcircled{\tiny 7}}
        }{}
    \end{overpic}
    % \includegraphics[width=.95\linewidth]{./imgs/MetaF2N-architecture.pdf}
    %\fbox{\rule{0pt}{2.4in} \rule{.9\linewidth}{0pt}}
    \caption{Pipeline of the proposed MetaF2N framework. During inference, the inner loop updates the initial parameter $\theta$ via a back-propagation step (dotted blue line), and the obtained $\theta_\mathit{n}$ is used to process the whole natural image in the outer loop. For training, the parameter update indeed relies on the gradients of outer loop loss $\mathcal{L}^\mathit{T_{i}}$ \wrt the initial parameter $\theta$ and the MaskNet parameter $\theta_\mathbf{m}$ (dotted red line). For easier understanding, the steps of the pipeline \iftoggle{usenumbers}{(remarked by circled numbers) }{}are introduced in \cref{sec:metaf2n_pipeline}. Please refer to \cref{alg:1,alg:2} for more details about the training and inference process.}
    \label{fig:metaf2n_pipeline}
    \vspace{-1em}
\end{figure*}

\section{Preliminary}
For efficiently utilizing the generalization ability of face restoration methods in blind image SR and achieving flexibility on image-specific degradations,
we have designed our MetaF2N using the model-agnostic meta-learning framework (MAML)~\cite{MAML}.
In this section, we briefly introduce the meta-learning framework based on MAML, which will be helpful for understanding our method.

Typically, a deep model $\mathit{f}$ is trained by a learning objective $\mathcal{L}$, which is directly dependent on the purpose of the task (\eg, $\ell_1$ loss for fidelity in image SR tasks), and the optimal parameter $\theta^\ast$ is obtained via
\begin{equation}
    \theta^\ast=\arg\min_{\theta}\mathcal{L}(\mathit{f}(\mathbf{x};\theta), \mathbf{y}),
    \label{eqn:metaf2n_dnn_objective}
\end{equation}
where $\mathbf{x}$ and $\mathbf{y}$ are input and ground-truth.
Intuitively, the objective of \cref{eqn:metaf2n_dnn_objective} is \textit{to get a $\theta$ leading to the lowest $\mathcal{L}$}.

Instead, MAML~\cite{MAML} aims to learn a parameter $\theta^\ast$, which can be efficiently fine-tuned toward the test sample with one or several steps.
Taking the one-step situation as an example, the learning objective can be written as
\begin{equation}
    \theta^\ast=\arg\min_{\theta}\sum_\mathit{T_\mathit{i}\sim\mathit{p(T)}}{\mathcal{L}^\mathit{T_{i}}(\mathit{f}(\mathbf{x}; \theta-\alpha\tfrac{\partial{\mathcal{L}^\mathit{T_{i}}_\mathit{in}}}{\partial\theta}), \mathbf{y})}\,,
    \label{eqn:metaf2n_maml_objective}
\end{equation}
where $\mathit{p(T)}$, $\mathcal{L}^\mathit{T_{i}}_\mathit{in}$, and $\alpha$ respectively represent the distribution (or collection) of tasks, the inner loop loss function for the task $\mathit{T_{i}}$, and learning rate for the inner loop.
Here, we can regard $\theta-\alpha\tfrac{\partial{\mathcal{L}^\mathit{T_{i}}_\mathit{in}}}{\partial\theta}$ as a whole. Then, similar to \cref{eqn:metaf2n_dnn_objective}, the objective of \cref{eqn:metaf2n_maml_objective} can be interpreted as \textit{to get a $\theta$ leading to the lowest $\mathcal{L}^\mathit{T_{i}}$ after one back-propagation step \wrt $\mathcal{L}^\mathit{T_{i}}_\mathit{in}$}.
In other words, a model trained under the MAML settings can adapt to new domains or tasks with a limited number of fine-tuning steps (\ie, $\theta-\alpha\tfrac{\partial{\mathcal{L}^\mathit{T_{i}}_\mathit{in}}}{\partial\theta}$).

\section{Method}
\label{sec:metaf2n_method}

In this work, we aim to obtain a blind image SR model $\mathit{f}$, which can be efficiently adapted to process the degradation of a real-world low-quality image $\mathbf{I}_\mathit{LR}$ under the guidance of its face regions (denoted by $\mathbf{I}^{\scriptsize\smiley{}}_\mathit{LR}$) and the faces processed by blind face restoration methods (denoted by $\mathbf{I}^{\scriptsize\smiley{}}_\mathit{BFR}$).

Considering the complex and wide-range degradations in real-world low-quality images, we take advantage of MAML to design our method for processing image-specific degradations.
Then \cref{eqn:metaf2n_maml_objective} can be rewritten as,
\begin{equation}
    \theta^\ast=\arg\min_{\theta}\sum_\mathit{T_\mathit{i}\sim\mathit{p(T)}}{\mathcal{L}^\mathit{T_{i}}(\mathit{f}(\mathbf{I}_\mathit{LR}; \theta-\alpha\tfrac{\partial{\mathcal{L}^\mathit{T_{i}}_\mathit{in}}}{\partial\theta}), \mathbf{I})},
    \label{eqn:metaf2n_blindsr_objective}
\end{equation}

where $\mathit{p(T)}$ is the distribution of different degradations, $\mathit{T_{i}}$ is the task for one specific degradation, $\theta$ is the parameter of $\mathit{f}$ and  $\mathbf{I}$ denotes the ground-truth image. The design of $\mathcal{L}^\mathit{T_{i}}_\mathit{in}$ will be given in \cref{sec:metaf2n_method_inner_loop}.

% \vspace{0.5em}
% \noindent\textbf{inner loop Design.}
\subsection{Inner Loop Design}
\label{sec:metaf2n_method_inner_loop}
One of the key {factors} of applying MAML~\cite{MAML} to blind image SR tasks is the design of $\mathcal{L}^\mathit{T_{i}}_\mathit{in}$ in \cref{eqn:metaf2n_blindsr_objective}.
Specifically, for adapting to the new degradation of an $\mathbf{I}_\mathit{LR}$, the data for calculating $\mathcal{L}^\mathit{T_{i}}_\mathit{in}$ should have the same distribution as $\mathbf{I}_\mathit{LR}$.
%For high-level tasks like (few-shot) image classification, one can provide labels for several samples easily, which serve as training data in the inner loop.
For MAML-based non-blind image SR methods~\cite{MZSR,MLSR}, one can construct inner loop data by further down-sampling $\mathbf{I}_\mathit{LR}$ with the known degradation $\mathit{d}$ to get $\mathbf{I}_\mathit{LLR}=\mathit{d}(\mathbf{I}_\mathit{LR})$.
Then assuming the inner loop loss function to be $\ell_1$, we can obtain $\mathcal{L}^\mathit{T_{i}}_\mathit{in}=\|\mathit{f}(\mathbf{I}_\mathit{LLR};\theta) - \mathbf{I}_\mathit{LR}\|_1$.
However, for blind image SR, it is almost impossible to get extra image pairs with the same degradation as $\mathbf{I}_\mathit{LR}$ since $\mathit{d}$ is unknown.

% $\mathbf{I}^{\frownie{}}_\mathit{LR}$

Inspired by ReDegNet~\cite{ReDegNet}, which shows that faces
%\lxm{share similar degradation as the whole image and
can be utilized to extract the degradation representation for the whole image, we directly construct the inner loop data with the face regions.
% In specific, the face regions of $\mathbf{I}_\mathit{LR}$ are denoted by $\mathbf{I}^{\iftoggle{useemoji}{\scriptsize\smiley{}}{\mathit{face}}}_\mathit{LR}$.
Thanks to the great generalization ability of blind face restoration methods~\cite{GFPGAN,GPEN,GLEAN,zhu2022blind}, we are able to obtain a pseudo ground-truth for the face regions, which is denoted by $\mathbf{I}^{\iftoggle{useemoji}{\scriptsize\smiley{}}{\mathit{face}}}_\mathit{BFR}=\mathit{f}_\mathit{BFR}(\mathbf{I}^{\iftoggle{useemoji}{\scriptsize\smiley{}}{\mathit{face}}}_\mathit{LR};\theta_\mathit{BFR})$, where $\mathit{f}_\mathit{BFR}$ is the blind face restoration model with pre-trained parameter $\theta_\mathit{BFR}$.
In this paper, we follow ReDegNet~\cite{ReDegNet} and adopt GPEN~\cite{GPEN} as $\mathit{f}_\mathit{BFR}$.
With the low-quality face regions $\mathbf{I}^{\iftoggle{useemoji}{\scriptsize\smiley{}}{\mathit{face}}}_\mathit{LR}$ and the recovered $\mathbf{I}^{\iftoggle{useemoji}{\scriptsize\smiley{}}{\mathit{face}}}_\mathit{BFR}$, the inner loop loss function can be defined by
\begin{equation}
    \mathcal{L}^\mathit{T_{i}}_\mathit{in} = \|\mathit{f}(\mathbf{I}^{\iftoggle{useemoji}{\scriptsize\smiley{}}{\mathit{face}}}_\mathit{LR};\theta)-\mathbf{I}^{\iftoggle{useemoji}{\scriptsize\smiley{}}{\mathit{face}}}_\mathit{BFR}\|_1.
    \label{eqn:metaf2n_inner_loop}
\end{equation}

% \vspace{0.5em}
% \noindent\textbf{Design of MaskNet.}
\subsection{Adaptive Loss Weighting with MaskNet}
Despite the appealing visual quality of recent blind face restoration methods, there inevitably exist gaps between generated faces $\mathbf{I}^{\iftoggle{useemoji}{\scriptsize\smiley{}}{\mathit{face}}}_\mathit{BFR}$ and ground-truth ones $\mathbf{I}^{\iftoggle{useemoji}{\scriptsize\smiley{}}{\mathit{face}}}$.
For inaccurate regions, the degradation either explicitly (ReDegNet~\cite{ReDegNet}) or implicitly (Our MetaF2N) described by the $(\mathbf{I}^{\iftoggle{useemoji}{\scriptsize\smiley{}}{\mathit{face}}}_\mathit{LR}, \mathbf{I}^{\iftoggle{useemoji}{\scriptsize\smiley{}}{\mathit{face}}}_\mathit{BFR})$ pair also deviate from that by $(\mathbf{I}^{\iftoggle{useemoji}{\scriptsize\smiley{}}{\mathit{face}}}_\mathit{LR}, \mathbf{I}^{\iftoggle{useemoji}{\scriptsize\smiley{}}{\mathit{face}}})$.

\begin{figure}[th]
    \centering
    \scriptsize
    \setlength\abovecaptionskip{0.5\baselineskip} 
    \setlength\belowcaptionskip{0.5\baselineskip}
    \if 0
    \begin{overpic}[grid=false,percent,width=.95\linewidth]{./imgs/MetaF2N-architecture-bak.pdf}
        \put(3.5 , 22  ){GPEN $\mathit{g}$}
        \put(13.9, 16  ){MaskNet $\mathit{f}_\mathbf{m}$}
        \put(15.2, 14.5){\cref{eqn:metaf2n_masknet}}
        %\put(2.4 , 8.2 ){SR Model $\mathit{f}$}
        \put(2.8 , 8.2 ){$\mathit{f}(\mathbf{I}^{\iftoggle{useemoji}{\scriptsize\smiley{}}{\mathit{face}}}_\mathit{LR};\theta)$}
        \put(27  , 8.2 ){$\mathcal{L}_\mathit{in}$}
        \put(5.5 , 2.6 ){$\theta$}
        \put(23.1 , 2.6 ){$\theta_\mathbf{n}\!=\!\theta\!-\!\alpha\tfrac{\partial\mathcal{L}_\mathit{in}}{\partial\theta}$}
        % \put(52.7, 13.8){SR Model $\mathit{f}$}
        \put(52.7, 13.8){$\mathit{f}(\mathbf{I}_\mathit{LR};\theta_\mathbf{n})$}
        \put(79.3, 13.8){$\mathcal{L}$}
    \end{overpic}
    \fi
    \begin{overpic}[grid=false,percent,width=.95\linewidth]{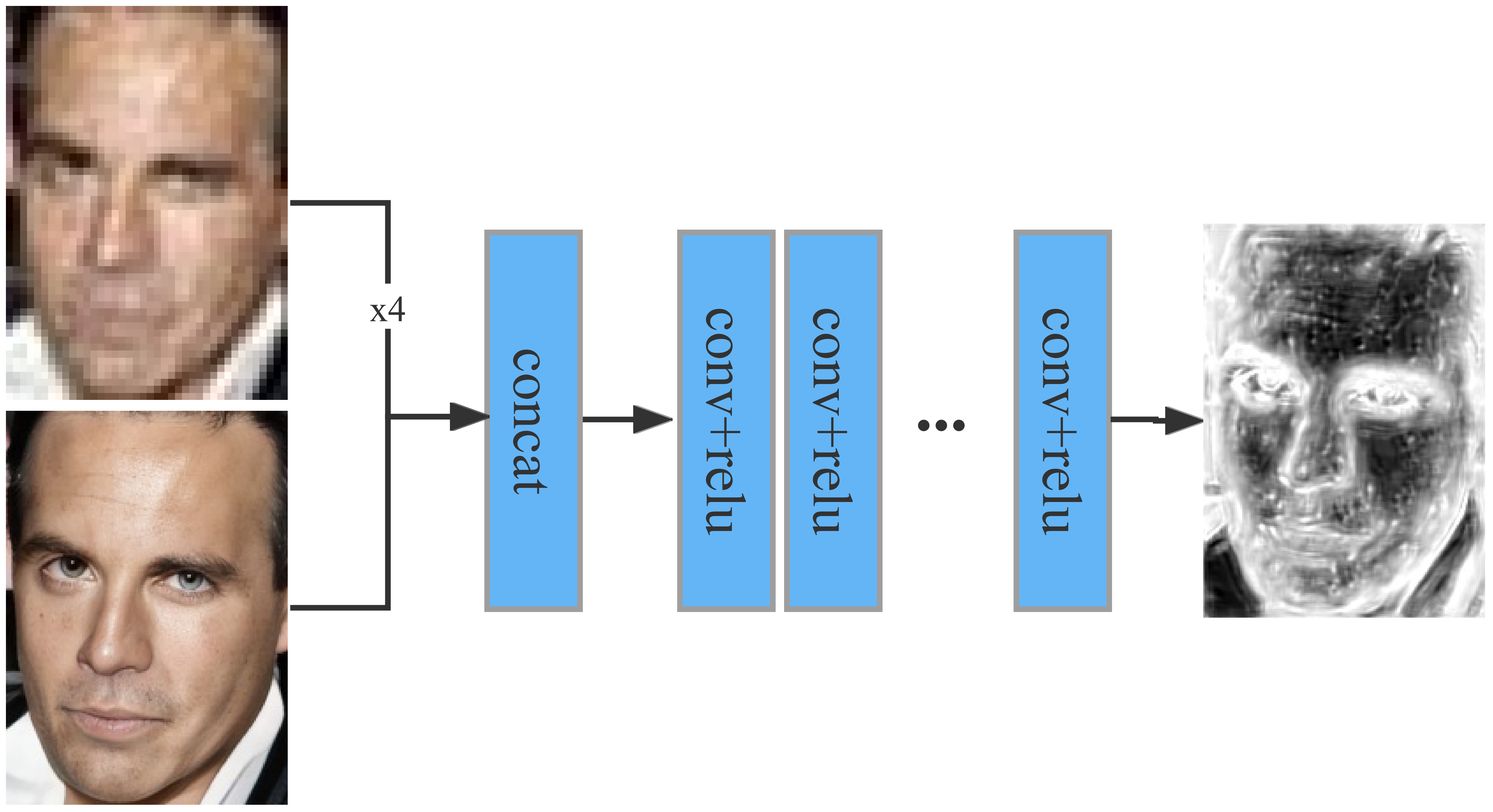}

        \put(1 , 50){\color{white}$\mathbf{I}^{\iftoggle{useemoji}{\scriptsize\smiley{}}{\mathit{face}}}_\mathit{LR}$}
        \put(1, 23){\color{white}$\mathbf{I}^{\iftoggle{useemoji}{\scriptsize\smiley{}}{\mathit{face}}}_\mathit{BFR}$}
        \put(82, 37){\color{black}$\mathbf{m}$}

    \end{overpic}

    \caption{The network structure of MaskNet.}
    \label{fig:masknet}
    \vspace{-1em}
\end{figure}
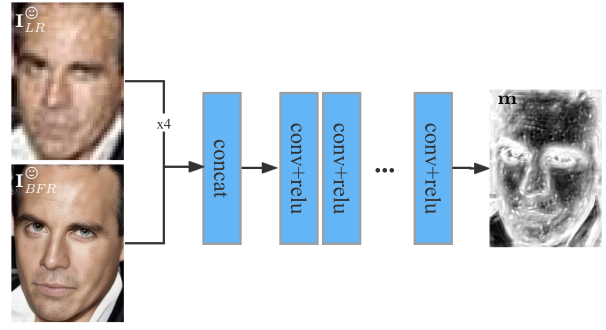

However, different regions of $\mathbf{I}^{\iftoggle{useemoji}{\scriptsize\smiley{}}{\mathit{face}}}_\mathit{BFR}$ are treated equally in \cref{eqn:metaf2n_inner_loop}, which will have negative effect on the final results.
As a remedy, we propose to predict loss weights adaptively for different regions via a MaskNet.
With the weight map $\mathbf{m}$ generated by the MaskNet, \cref{eqn:metaf2n_inner_loop} can be rewritten as
\begin{equation}
    \mathcal{L}^\mathit{T_{i}}_\mathit{in} = \|\mathbf{m}\cdot(\mathit{f}(\mathbf{I}^{\iftoggle{useemoji}{\scriptsize\smiley{}}{\mathit{face}}}_\mathit{LR};\theta)-\mathbf{I}^{\iftoggle{useemoji}{\scriptsize\smiley{}}{\mathit{face}}}_\mathit{BFR})\|_1.
    \label{eqn:metaf2n_inner_loop_with_masknet}
\end{equation}
Ideally, $\mathbf{m}$ should be generated from the $(\mathbf{I}^{\iftoggle{useemoji}{\scriptsize\smiley{}}{\mathit{face}}}_\mathit{BFR}, \mathbf{I}^{\iftoggle{useemoji}{\scriptsize\smiley{}}{\mathit{face}}})$ pair.
Unfortunately, $\mathbf{I}$ is unavailable for test samples, a possible solution is solely predicting from $\mathbf{I}^{\iftoggle{useemoji}{\scriptsize\smiley{}}{\mathit{face}}}_\mathit{BFR}$, yet the improvement is marginal.
Analogous to degraded-reference IQA~\cite{DRIQA_ICCV21,DRIQA_TIP23} which predicts IQA metrics from (input, output) pairs rather than (output, GT) pairs in full-reference IQA, we predict $\mathbf{m}$ from the $(\mathbf{I}^{\iftoggle{useemoji}{\scriptsize\smiley{}}{\mathit{face}}}_\mathit{LR}, \mathbf{I}^{\iftoggle{useemoji}{\scriptsize\smiley{}}{\mathit{face}}}_\mathit{BFR})$ pair, \ie,
\begin{equation}
    \mathbf{m}=\mathit{f}_\mathbf{m}(\mathbf{I}^{\iftoggle{useemoji}{\scriptsize\smiley{}}{\mathit{face}}}_\mathit{LR}, \mathbf{I}^{\iftoggle{useemoji}{\scriptsize\smiley{}}{\mathit{face}}}_\mathit{BFR}; \theta_\mathbf{m}),
    \label{eqn:metaf2n_masknet}
\end{equation}
where $\mathit{f}_\mathbf{m}$ is the MaskNet, which is a simple network composed of 8 convolution layers shown in \cref{fig:masknet}.
As further shown in the ablation studies, the degraded-reference solution in \cref{eqn:metaf2n_masknet} performs better than solely predicting from $\mathbf{I}^{\iftoggle{useemoji}{\scriptsize\smiley{}}{\mathit{face}}}_\mathit{BFR}$.

\begin{algorithm}[t]
        \DontPrintSemicolon
        \SetAlgoLined
        % \small
        \KwInput{{Distribution of different degradation restoration tasks $\mathit{p(T)}$}; \\
        \hspace*{0.39in} High-quality natural image dataset $\mathcal{D}_\mathit{HR}$; \\
        \hspace*{0.39in} High-quality face dataset $\mathcal{{D}_{\iftoggle{useemoji}{\scriptsize\smiley{}}{\mathit{face}}}}$; \\ 
        \hspace*{0.39in} Learning rates $\alpha, \beta, \gamma, \eta$;}
        
        \KwOutput{Model parameter $\theta, \theta_\mathbf{m}$}
        Initialize $\mathit{f}$, $\mathit{D}$ with $\theta$, $\theta_\mathit{D}$ from Real-ESRGAN~\cite{Real-ESRGAN} \\
        Initialize $\mathit{f}_\mathit{BFR}$ with $\theta_\mathit{BFR}$ from GPEN~\cite{GPEN} \\
        Random initialize $\mathit{f}_\mathbf{m}$ with $\theta_\mathbf{m}$ \\
        \For{all training steps}{
            {Sample batch of tasks $\mathit{T_{i}}\sim\mathit{p(T)}$}\\
            \For{all $\mathit{T_{i}}$}{
            Sample images $\mathbf{I}, \mathbf{I}^{\iftoggle{useemoji}{\scriptsize\smiley{}}{\mathit{face}}}$ from $\mathcal{{D}_{HR}}$ and $\mathcal{{D}_{\iftoggle{useemoji}{\scriptsize\smiley{}}{\mathit{face}}}}$ \\
            Create $\mathbf{I}, \mathbf{I}_\mathit{LR}, \mathbf{I}^{\iftoggle{useemoji}{\scriptsize\smiley{}}{\mathit{face}}}_\mathit{LR}$ of $\mathit{T_{i}}$ \\
            % \For{all samples in the batch}{
                $\mathbf{I}^{\iftoggle{useemoji}{\scriptsize\smiley{}}{\mathit{face}}}_\mathit{BFR} = \mathit{f}_\mathit{BFR}(\mathbf{I}^{\iftoggle{useemoji}{\scriptsize\smiley{}}{\mathit{face}}}_\mathit{LR};\theta_\mathit{BFR})$ \\
                $\mathbf{m} = \mathit{f}_\mathbf{m}(\mathbf{I}^{\iftoggle{useemoji}{\scriptsize\smiley{}}{\mathit{face}}}_\mathit{LR}, \mathbf{I}^{\iftoggle{useemoji}{\scriptsize\smiley{}}{\mathit{face}}}_\mathit{BFR}; \theta_\mathbf{m})$ \\

                \For{$\mathrm{1}$ step of inner loop}{
                     $\mathcal{L}^\mathit{T_{i}}_\mathit{in} = \|\mathbf{m}\cdot(\mathit{f}(\mathbf{I}^{\iftoggle{useemoji}{\scriptsize\smiley{}}{\mathit{face}}}_\mathit{LR};\theta)-\mathbf{I}^{\iftoggle{useemoji}{\scriptsize\smiley{}}{\mathit{face}}}_\mathit{BFR})\|_1$ \\
                     $\theta_\mathbf{n} \leftarrow {\theta - \alpha \nabla_\mathit{\theta}{\mathcal{L}^\mathit{T_{i}}_\mathit{in}}}(\theta)$ \\
                }
                $\mathbf{I}_\mathit{SR} = \mathit{f}(\mathbf{I}_\mathit{LR};\theta_\mathbf{n})$ \\
          
                $\mathcal{L}^\mathit{T_{i}}$ is defined as \cref{eqn:metaf2n_learning_objective} \\
                $\mathcal{L}^\mathit{T_{i}}_\mathit{D}$ is defined as \cref{eqn:metaf2n_loss_D} \\
            }

            $\theta \leftarrow \theta - \beta \nabla_\theta\sum_\mathit{T_\mathit{i}\sim\mathit{p(T)}}\mathcal{L}^\mathit{T_{i}}$  \\
            $\theta_\mathbf{m} \leftarrow \theta_\mathbf{m} - \gamma \nabla_{\theta_\mathbf{m}}\sum_\mathit{T_\mathit{i}\sim\mathit{p(T)}}\mathcal{L}^\mathit{T_{i}}$  \\
            $\theta_\mathit{D} \leftarrow \theta_\mathit{D} - \eta \nabla_{\theta_\mathit{D}}\sum_\mathit{T_\mathit{i}\sim\mathit{p(T)}}\mathcal{L}^\mathit{T_{i}}_\mathit{D}$  \\
            
        }
        \caption{Training Process of MetaF2N}
        \label{alg:1}
        
\end{algorithm}

\begin{algorithm}[t]
        \DontPrintSemicolon
        \SetAlgoLined
        \KwInput{LR test image $\mathbf{I}_{LR}$; \\
        \hspace*{0.39in} Trained model parameter $\theta, \theta_\mathbf{m}$; \\
        \hspace*{0.39in} Pre-trained GPEN~\cite{GPEN} parameter $\theta_\mathit{BFR}$; \\
        \hspace*{0.39in} Fine-tuning steps $\mathit{n}$ and learning rate $\alpha$;}
        \KwOutput{Super-resolved image $\mathbf{I}_{SR}$}
        Initialize $\mathit{f}$, $\mathit{f}_\mathbf{m}$ with $\theta$, $\theta_\mathbf{m}$ \\
        Initialize $\mathit{f}_\mathit{BFR}$ with $\theta_\mathit{BFR}$ from GPEN~\cite{GPEN} \\
        Extract face regions $\mathbf{I}^{\iftoggle{useemoji}{\scriptsize\smiley{}}{\mathit{face}}}_\mathit{LR}$ from $\mathbf{I}_\mathit{LR}$ \\

        $\mathbf{I}^{\iftoggle{useemoji}{\scriptsize\smiley{}}{\mathit{face}}}_\mathit{BFR} = \mathit{f}_\mathit{BFR}(\mathbf{I}^{\iftoggle{useemoji}{\scriptsize\smiley{}}{\mathit{face}}}_\mathit{LR};\theta_\mathit{BFR})$ \\
%        Randomly crop patches from $(\mathbf{I}^{\iftoggle{useemoji}{\scriptsize\smiley{}}{\mathit{face}}}_\mathit{LR}, \mathbf{I}^{\iftoggle{useemoji}{\scriptsize\smiley{}}{\mathit{face}}}_\mathit{BFR})$ \\
        $\mathbf{m} = \mathit{f}_\mathbf{m}(\mathbf{I}^{\iftoggle{useemoji}{\scriptsize\smiley{}}{\mathit{face}}}_\mathit{LR}, \mathbf{I}^{\iftoggle{useemoji}{\scriptsize\smiley{}}{\mathit{face}}}_\mathit{BFR}; \theta_\mathbf{m})$ \\
%        $\theta_\mathbf{n} \leftarrow \theta$ \\
        \For{n steps of inner loop}
        {
            $\mathcal{L}_\mathit{in} = \|\mathbf{m}\cdot(\mathit{f}(\mathbf{I}^{\iftoggle{useemoji}{\scriptsize\smiley{}}{\mathit{face}}}_\mathit{LR};\theta)-\mathbf{I}^{\iftoggle{useemoji}{\scriptsize\smiley{}}{\mathit{face}}}_\mathit{BFR})\|_1$ \\
            % $\theta_\mathbf{n} \leftarrow {\theta - \alpha \nabla_\mathit{\theta}{\mathcal{L}_\mathit{in}}}(\theta_\mathbf{n})$ \\
            $\theta \leftarrow {\theta - \alpha \nabla_\mathit{\theta}{\mathcal{L}_\mathit{in}}}(\theta)$ \\
        }
        $\theta_\mathbf{n} \leftarrow \theta$ \\
        $\mathbf{I}_\mathit{SR} = \mathit{f}(\mathbf{I}_\mathit{LR};\theta_\mathbf{n})$ \\
        \Return  $\mathbf{I}_\mathit{SR}$ \\

        \caption{Inference Process of MetaF2N}
        \label{alg:2}
\end{algorithm}

% \vspace{0.5em}
% \noindent\textbf{Overall Pipeline.}
\subsection{Overall Pipeline and Learning Objective}
\label{sec:metaf2n_pipeline}
With the inner loop designed in \cref{eqn:metaf2n_inner_loop_with_masknet}, we can build the pipeline as shown in \cref{fig:metaf2n_pipeline}.
In specific, given a low-quality image $\mathbf{I}_\mathit{LR}$ containing face regions denoted by $\mathbf{I}^{\iftoggle{useemoji}{\scriptsize\smiley{}}{\mathit{face}}}_\mathit{LR}$, we pass $\mathbf{I}^{\iftoggle{useemoji}{\scriptsize\smiley{}}{\mathit{face}}}_\mathit{LR}$ through the SR network $\mathit{f}$ and obtain the inner loop result $\mathbf{I}^{\iftoggle{useemoji}{\scriptsize\smiley{}}{\mathit{face}}}_\mathit{SR}$\iftoggle{usenumbers}{$^{\small\textcircled{\scriptsize 1}}$}{}.
For a fair comparison against existing methods, we take the model of Real-ESRGAN~\cite{ESRGAN} as the SR network, whose parameter is used to initialize the $\theta$ in the inner loop.
To construct supervision for the inner loop, we use a pre-trained GPEN~\cite{GPEN} model, whose parameter is kept fixed in our whole pipeline\iftoggle{usenumbers}{$^{\small\textcircled{\scriptsize 2}}$}{}.
Then we can obtain the weight map $\mathbf{m}$ following \cref{eqn:metaf2n_masknet}\iftoggle{usenumbers}{$^{\small\textcircled{\scriptsize 3}}$}{}, as well as the inner loop loss $\mathcal{L}^\mathit{T_{i}}_\mathit{in}$ following \cref{eqn:metaf2n_inner_loop_with_masknet}\iftoggle{usenumbers}{$^{\small\textcircled{\scriptsize 4}}$}{}, which can be used to generate a temporary parameter $\theta_\mathbf{n}=\theta-\alpha\tfrac{\partial\mathcal{L}^\mathit{T_{i}}_\mathit{in}}{\partial\theta}$\iftoggle{usenumbers}{$^{\small\textcircled{\scriptsize 5}}$}{} for processing the whole natural image in the outer loop, \ie,
\begin{equation}
    \mathbf{I}_\mathit{SR}=\mathit{f}(\mathbf{I}_\mathit{LR}; \theta_\mathbf{n})=\mathit{f}(\mathbf{I}_\mathit{LR}; \theta-\alpha\tfrac{\partial\mathcal{L}^\mathit{T_{i}}_\mathit{in}}{\partial\theta})\iftoggle{usenumbers}{^{\small\textcircled{\scriptsize 6}}}{}.
\end{equation}

The above process\iftoggle{usenumbers}{ (as remarked by the circled numbers)}{} describes the pipeline of MetaF2N during inference and the first six steps in the training phase.

For training MetaF2N, we also need to calculate the outer loop loss\iftoggle{usenumbers}{$^{\small\textcircled{\scriptsize 7}}$}{}, which is composed of fidelity loss ($\ell_1$), LPIPS loss~\cite{LPIPS}, and GAN loss~\cite{GAN}.
In specific, the fidelity loss is
\begin{equation}
    \mathcal{L}_1=\|\mathbf{I}_\mathit{SR}-\mathbf{I}\|_1,
\end{equation}
and the LPIPS loss is defined by
\begin{equation}
    \mathcal{L}_\mathrm{LPIPS} = \|\phi(\mathbf{I}_\mathit{SR})-\phi(\mathbf{I})\|_2,
\end{equation}
where $\phi$ is the pre-trained AlexNet~\cite{AlexNet} feature extractor for calculating LPIPS.
The adversarial loss follows the setting of Real-ESRGAN~\cite{Real-ESRGAN}, which is defined by,
\begin{equation}
    \mathcal{L}_\mathrm{adv} = -\mathbb{E}[\log(\mathit{D}(\mathbf{I}_\mathit{SR}))],
\end{equation}
where the discriminator $\mathit{D}$ is iteratively trained along with the SR network, \ie,
\begin{equation}
    \mathcal{L}_\mathit{D} = -\mathbb{E}[\log(\mathit{D}(\mathbf{I}))-\log(1-\mathit{D}(\mathbf{I}_\mathit{SR}))].
    \label{eqn:metaf2n_loss_D}
\end{equation}
Note that some works~\cite{DAWSON, Meta-KernelGAN} have explored training GANs under the meta-learning structure, and all of them update the discriminator loss in both the inner loop and outer loop.
However, to accelerate the training process and save memory, we only update the discriminator loss in the outer loop, which also shows satisfactory results.

To improve the numerical stability and avoid gradient exploding/vanishing, we constrain the MaskNet $\mathit{f}_\mathbf{m}$ via a regularization term, which is defined by
\begin{equation}
    \mathcal{L}_\mathrm{reg}=\|\mathbf{m}-\mathbf{1}\|_2.
\end{equation}
In summary, the learning objective of our MetaF2N (\ie, the outer loop loss) is
\begin{equation}
    \mathcal{L}^\mathit{T_{i}} = \lambda_1\mathcal{L}_1 + \lambda_2\mathcal{L}_\mathrm{LPIPS} + \lambda_3\mathcal{L}_\mathrm{adv} + \lambda_4\mathcal{L}_\mathrm{reg},
    \label{eqn:metaf2n_learning_objective}
\end{equation}
where the hyperparameters $\lambda_1$, $\lambda_2$, $\lambda_3$, and $\lambda_4$ are empirically set to 1, 0.5, 0.1, and 0.002, respectively.
Please refer to \cref{alg:1,alg:2} for more details on the training and inference process of the proposed MetaF2N.

\section{Experiments}

\begin{figure*}
	\centering
        \footnotesize
        \setlength\abovecaptionskip{0.2\baselineskip}
        \setlength\belowcaptionskip{0.2\baselineskip}

 		\begin{minipage}{0.13\linewidth}
		\centering
		\includegraphics[width=1.0\linewidth]{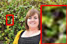}
	\end{minipage}
 	\begin{minipage}{0.13\linewidth}
		\centering
		\includegraphics[width=1.0\linewidth]{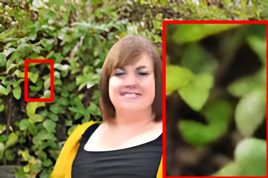}
	\end{minipage}
 	\begin{minipage}{0.13\linewidth}
		\centering
		\includegraphics[width=1.0\linewidth]{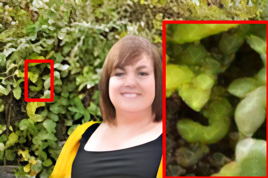}
	\end{minipage}
	\begin{minipage}{0.13\linewidth}
		\centering
		\includegraphics[width=1.0\linewidth]{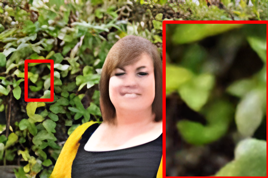}
	\end{minipage}
 	\begin{minipage}{0.13\linewidth}
		\centering
		\includegraphics[width=1.0\linewidth]{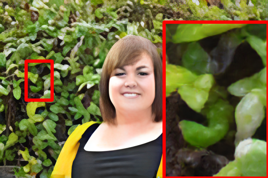}
	\end{minipage}
 	\begin{minipage}{0.13\linewidth}
		\centering
		\includegraphics[width=1.0\linewidth]{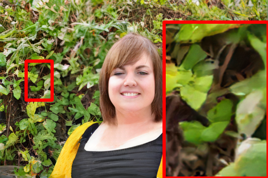}
	\end{minipage}
  	\begin{minipage}{0.13\linewidth}
		\centering
		\includegraphics[width=1.0\linewidth]{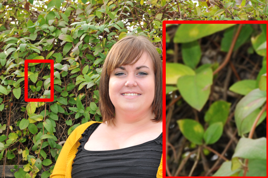}
	\end{minipage}

		\begin{minipage}{0.13\linewidth}
		\centering
		\includegraphics[width=1.0\linewidth]{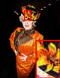}
	\end{minipage}
 	\begin{minipage}{0.13\linewidth}
		\centering
		\includegraphics[width=1.0\linewidth]{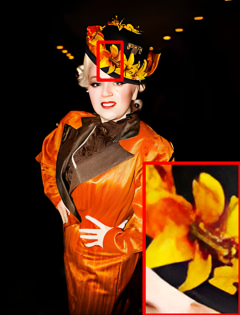}
	\end{minipage}
 	\begin{minipage}{0.13\linewidth}
		\centering
		\includegraphics[width=1.0\linewidth]{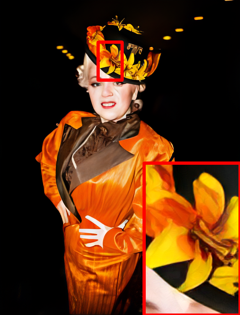}
	\end{minipage}
	\begin{minipage}{0.13\linewidth}
		\centering
		\includegraphics[width=1.0\linewidth]{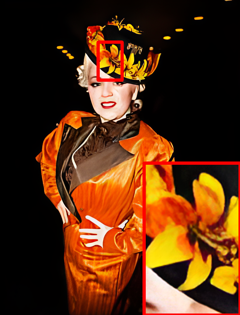}
	\end{minipage}
 	\begin{minipage}{0.13\linewidth}
		\centering
		\includegraphics[width=1.0\linewidth]{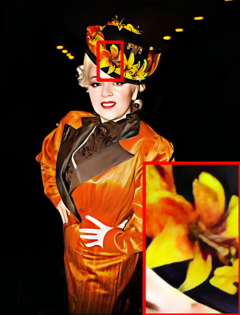}
	\end{minipage}
 	\begin{minipage}{0.13\linewidth}
		\centering
		\includegraphics[width=1.0\linewidth]{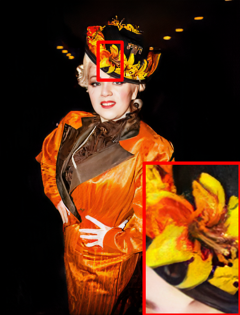}
	\end{minipage}
  	\begin{minipage}{0.13\linewidth}
		\centering
		\includegraphics[width=1.0\linewidth]{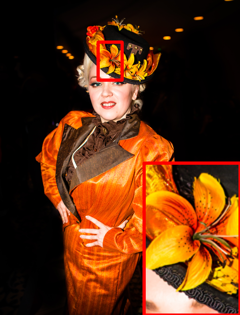}
	\end{minipage}

 	\begin{minipage}{0.13\linewidth}
		\centering
		\includegraphics[width=1.0\linewidth]{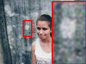}
	\end{minipage}
 	\begin{minipage}{0.13\linewidth}
		\centering
		\includegraphics[width=1.0\linewidth]{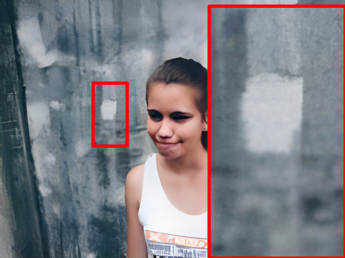}
	\end{minipage}
 	\begin{minipage}{0.13\linewidth}
		\centering
		\includegraphics[width=1.0\linewidth]{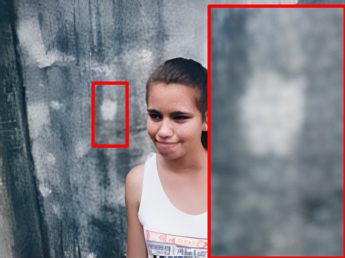}
	\end{minipage}
	\begin{minipage}{0.13\linewidth}
		\centering
		\includegraphics[width=1.0\linewidth]{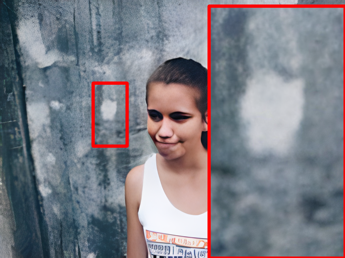}
	\end{minipage}
 	\begin{minipage}{0.13\linewidth}
		\centering
		\includegraphics[width=1.0\linewidth]{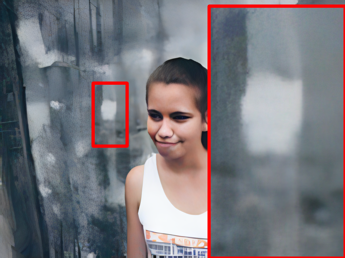}
	\end{minipage}
 	\begin{minipage}{0.13\linewidth}
		\centering
		\includegraphics[width=1.0\linewidth]{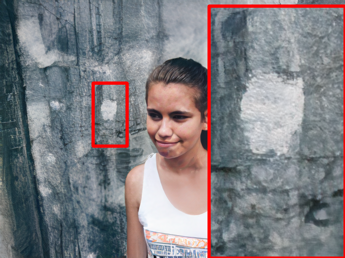}
	\end{minipage}
  	\begin{minipage}{0.13\linewidth}
		\centering
		\includegraphics[width=1.0\linewidth]{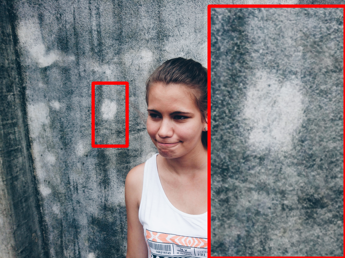}
	\end{minipage}

    \begin{minipage}{0.13\linewidth}
        \centering
        Synthetic LR
    \end{minipage}
    \begin{minipage}{0.13\linewidth}
        \centering
        Real-ESRGAN~\cite{Real-ESRGAN}
    \end{minipage}
    \begin{minipage}{0.13\linewidth}
        \centering
        BSRGAN~\cite{BSRGAN}
    \end{minipage}
    \begin{minipage}{0.13\linewidth}
        \centering
        MM-RealSR~\cite{MM-RealSR}
    \end{minipage}
    \begin{minipage}{0.13\linewidth}
        \centering
        ReDegNet~\cite{ReDegNet}
    \end{minipage}
    \begin{minipage}{0.13\linewidth}
        \centering
        Ours
    \end{minipage}
    \begin{minipage}{0.13\linewidth}
        \centering
        Ground-truth
    \end{minipage}
    
     \caption{Visual comparison against state-of-the-art blind SR methods on synthetic datasets. Note that the results of Ours are produced by MetaF2N with one-step fine-tuning. Please zoom in for better observation.}
    \label{fig:metaf2n_results_synthetic}
    \vspace{-1em}
\end{figure*}

\begin{figure*}
	\centering
        \footnotesize
        \setlength\abovecaptionskip{0.2\baselineskip}
        \setlength\belowcaptionskip{0.2\baselineskip}
	\begin{minipage}{0.13\linewidth}
		\centering
		\includegraphics[width=1.0\linewidth]{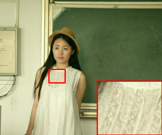}
	\end{minipage}
 	\begin{minipage}{0.13\linewidth}
		\centering
		\includegraphics[width=1.0\linewidth]{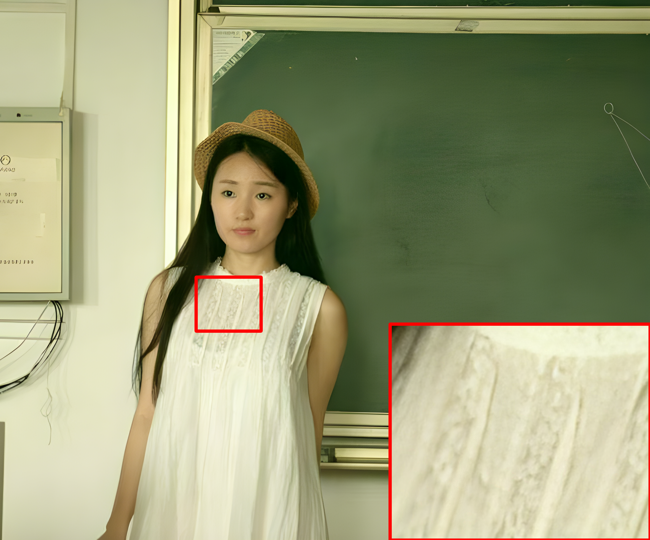}
	\end{minipage}
 	\begin{minipage}{0.13\linewidth}
		\centering
		\includegraphics[width=1.0\linewidth]{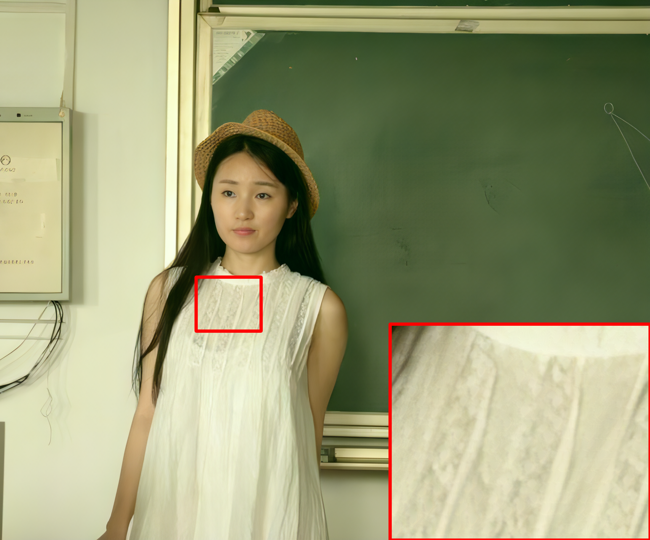}
	\end{minipage}
	\begin{minipage}{0.13\linewidth}
		\centering
		\includegraphics[width=1.0\linewidth]{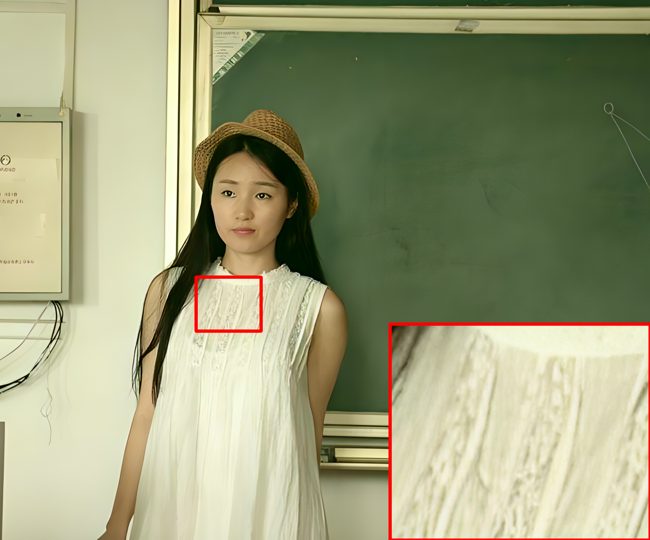}
	\end{minipage}
 	\begin{minipage}{0.13\linewidth}
		\centering
		\includegraphics[width=1.0\linewidth]{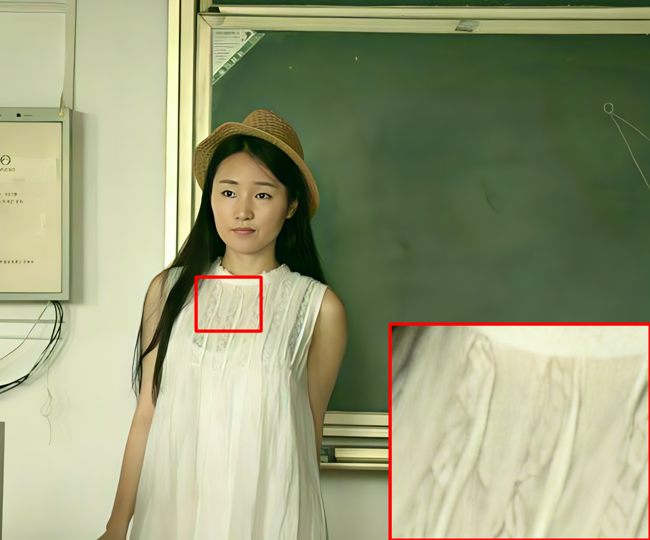}
	\end{minipage}
 	\begin{minipage}{0.13\linewidth}
		\centering
		\includegraphics[width=1.0\linewidth]{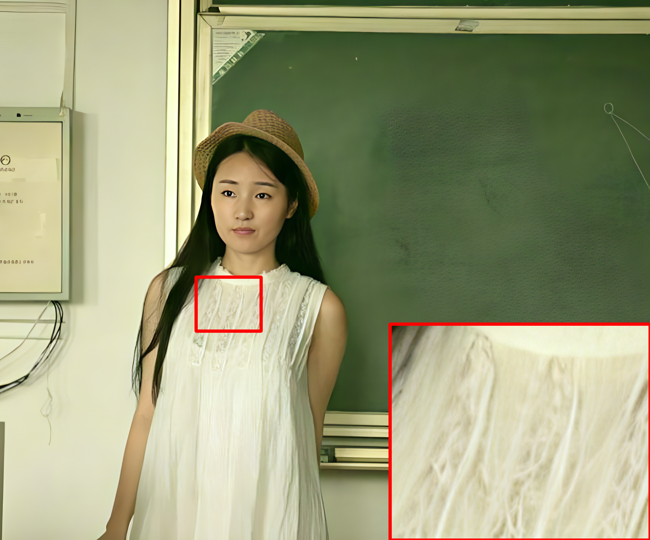}
	\end{minipage}
 	\begin{minipage}{0.13\linewidth}
		\centering
		\includegraphics[width=1.0\linewidth]{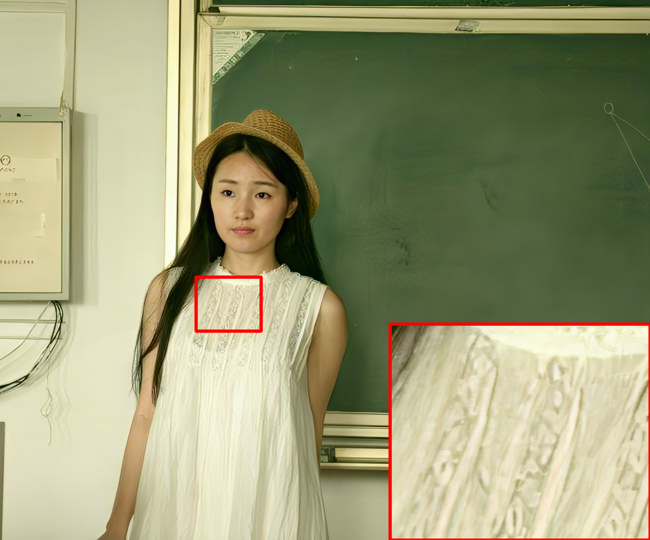}
	\end{minipage}

 	\begin{minipage}{0.13\linewidth}
		\centering
		\includegraphics[width=1.0\linewidth]{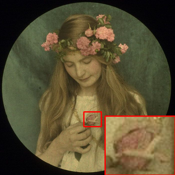}
	\end{minipage}
 	\begin{minipage}{0.13\linewidth}
		\centering
		\includegraphics[width=1.0\linewidth]{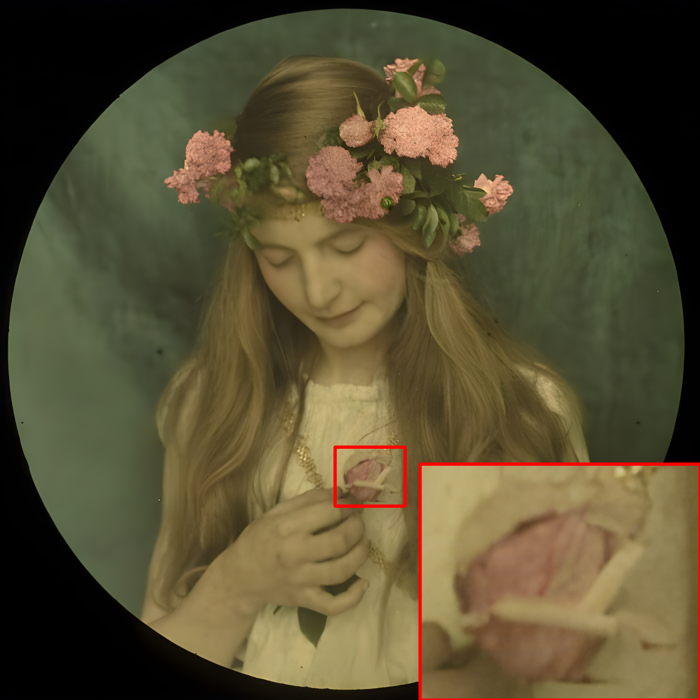}
	\end{minipage}
 	\begin{minipage}{0.13\linewidth}
		\centering
		\includegraphics[width=1.0\linewidth]{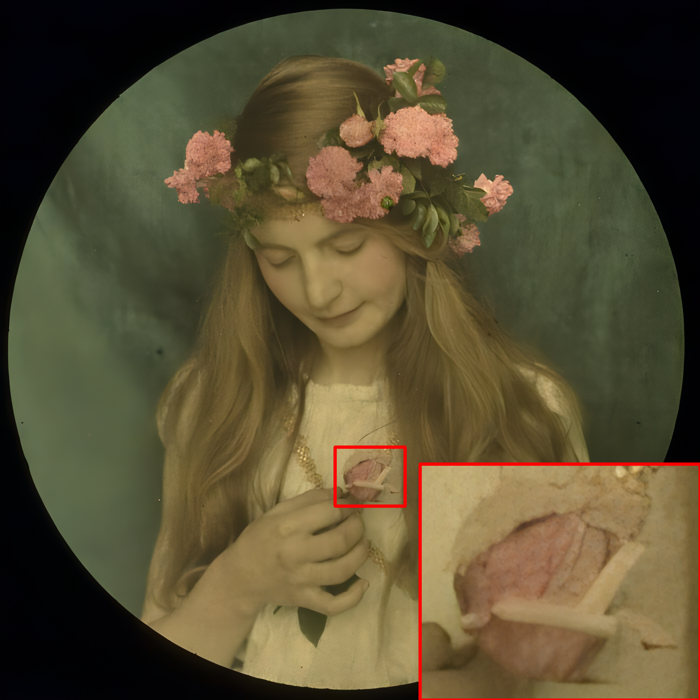}
	\end{minipage}
	\begin{minipage}{0.13\linewidth}
		\centering
		\includegraphics[width=1.0\linewidth]{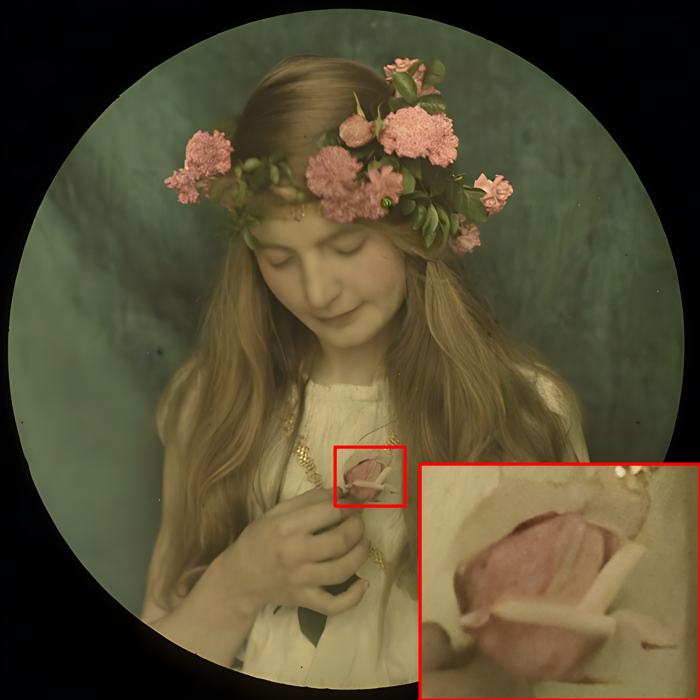}
	\end{minipage}
 	\begin{minipage}{0.13\linewidth}
		\centering
		\includegraphics[width=1.0\linewidth]{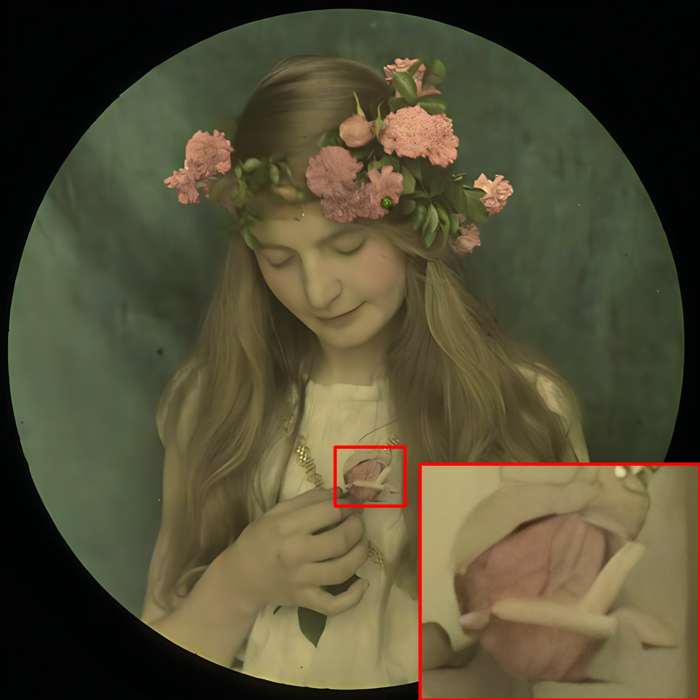}
	\end{minipage}
 	\begin{minipage}{0.13\linewidth}
		\centering
		\includegraphics[width=1.0\linewidth]{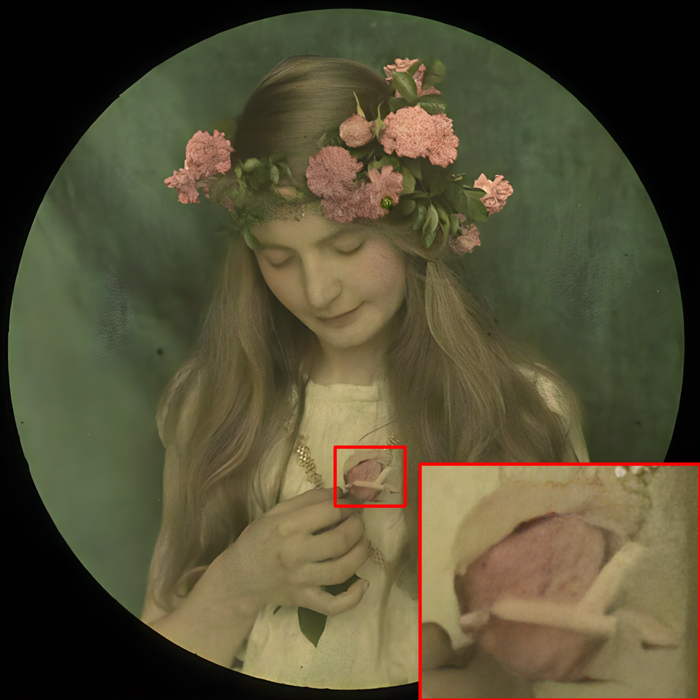}
	\end{minipage}
 	\begin{minipage}{0.13\linewidth}
		\centering
		\includegraphics[width=1.0\linewidth]{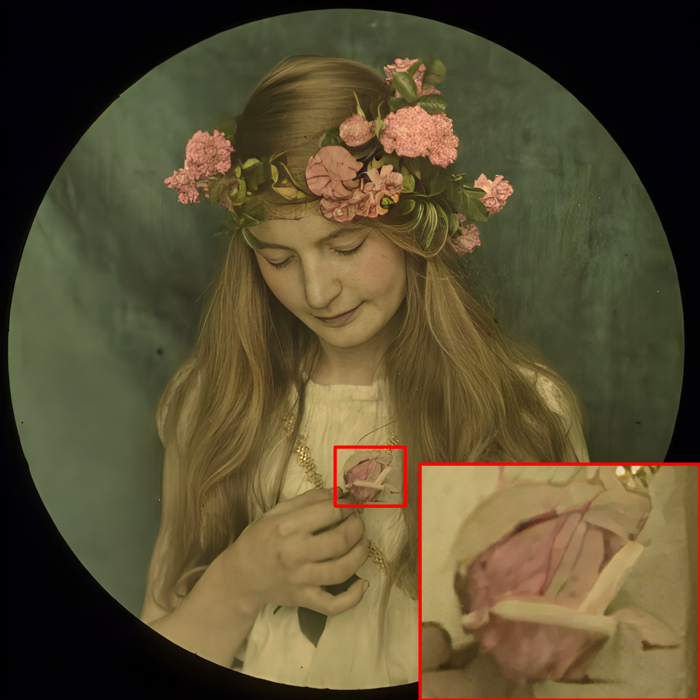}
	\end{minipage}

	\begin{minipage}{0.13\linewidth}
		\centering
		\includegraphics[width=1.0\linewidth]{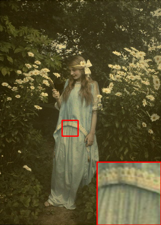}
	\end{minipage}
 	\begin{minipage}{0.13\linewidth}
		\centering
		\includegraphics[width=1.0\linewidth]{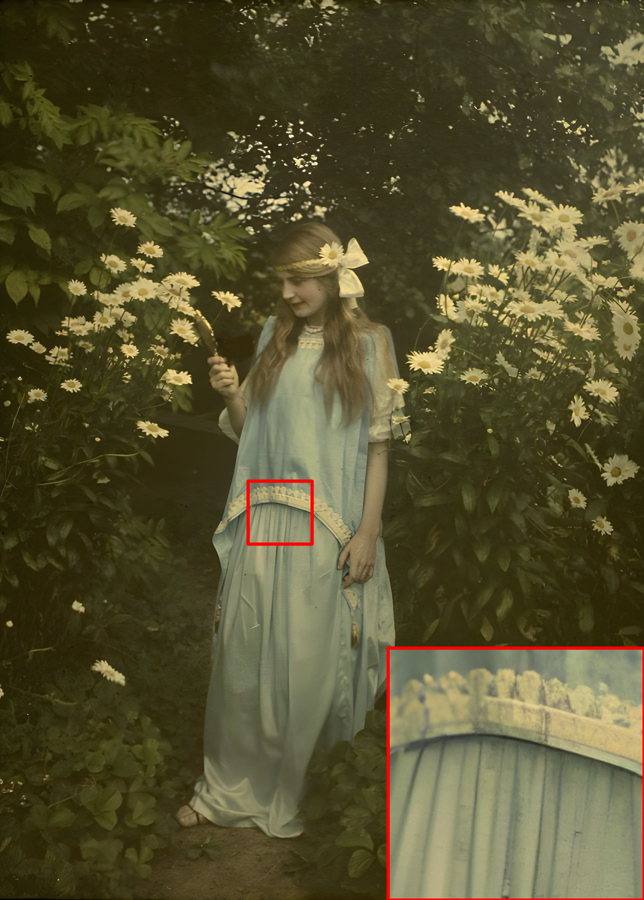}
	\end{minipage}
 	\begin{minipage}{0.13\linewidth}
		\centering
		\includegraphics[width=1.0\linewidth]{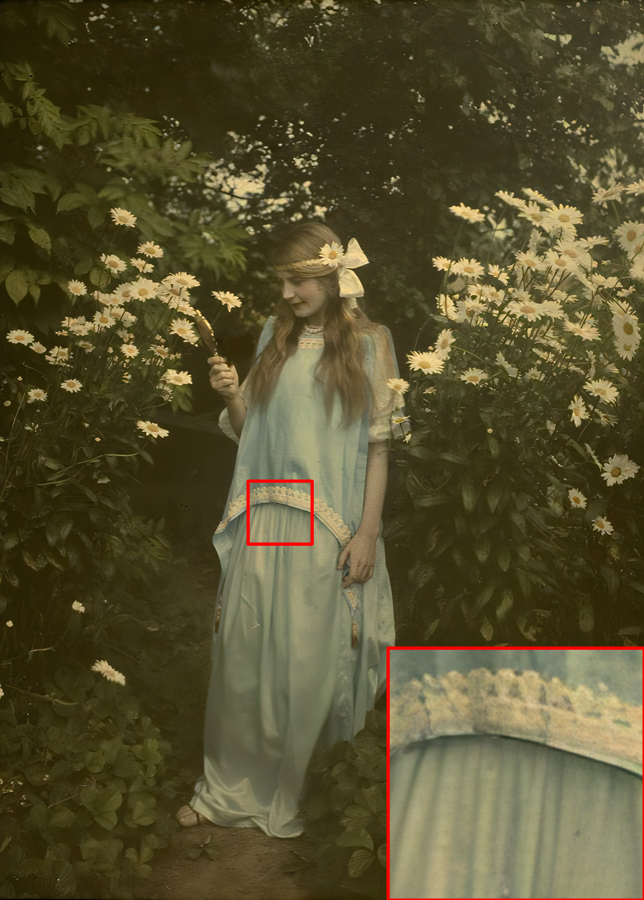}
	\end{minipage}
	\begin{minipage}{0.13\linewidth}
		\centering
		\includegraphics[width=1.0\linewidth]{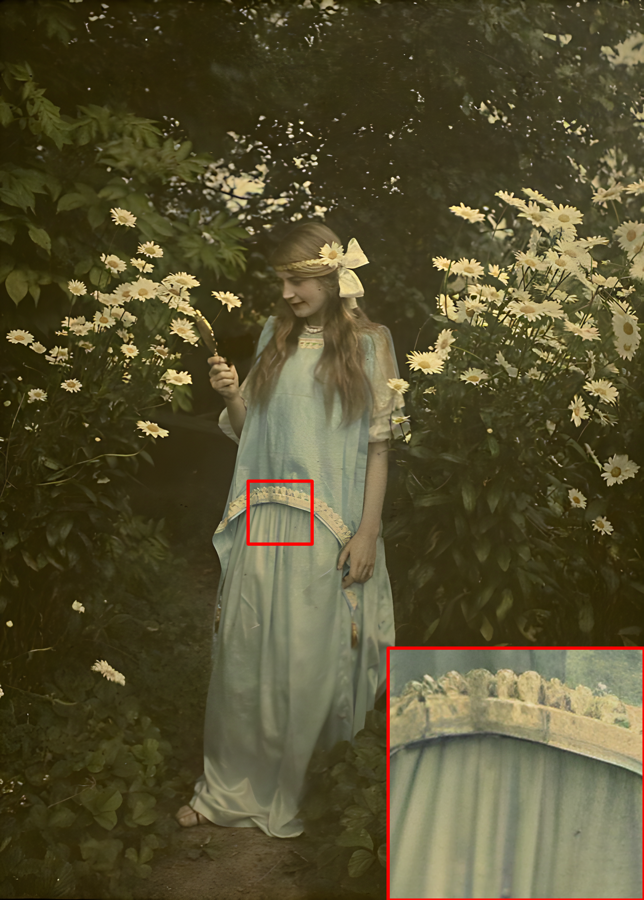}
	\end{minipage}
 	\begin{minipage}{0.13\linewidth}
		\centering
		\includegraphics[width=1.0\linewidth]{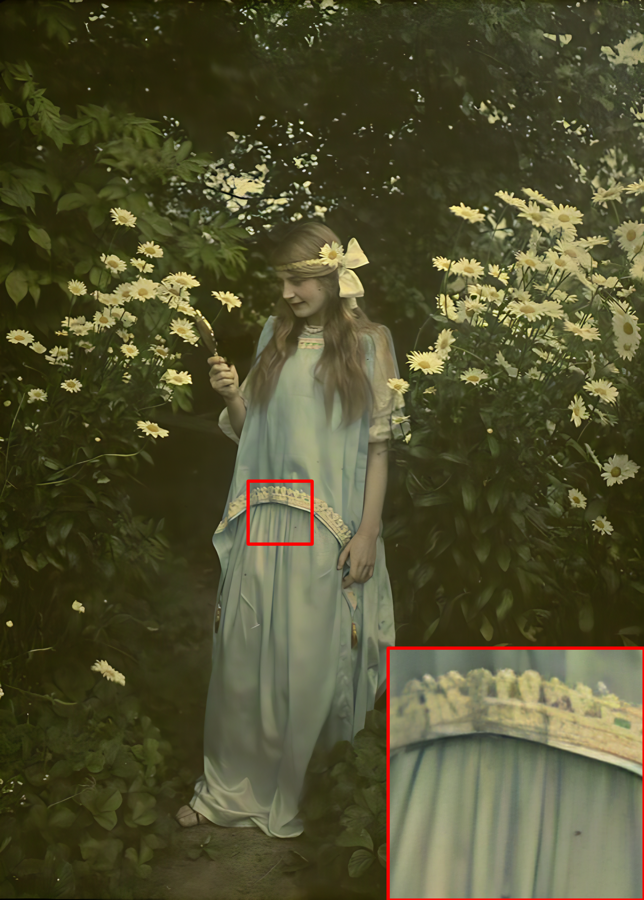}
	\end{minipage}
 	\begin{minipage}{0.13\linewidth}
		\centering
		\includegraphics[width=1.0\linewidth]{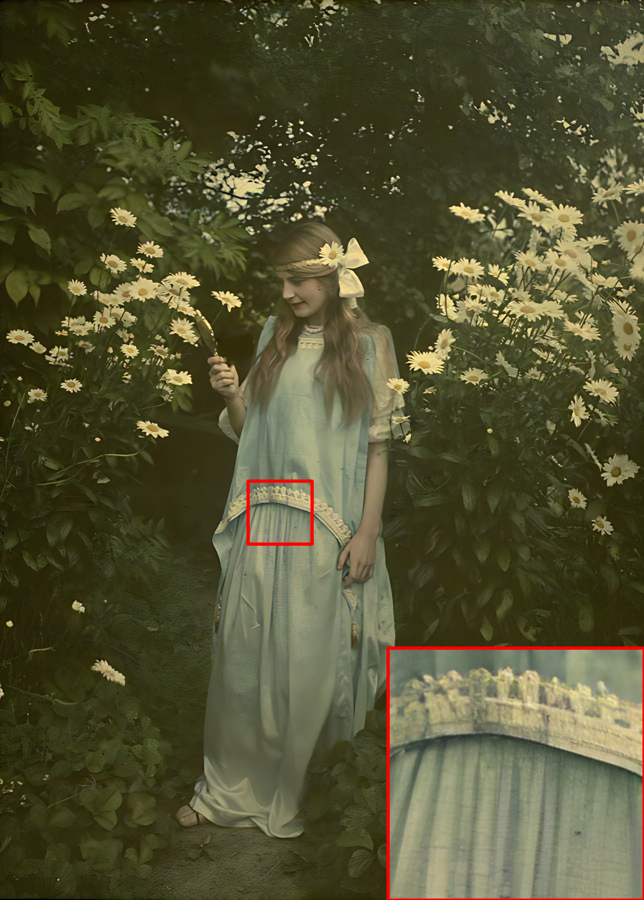}
	\end{minipage}
 	\begin{minipage}{0.13\linewidth}
		\centering
		\includegraphics[width=1.0\linewidth]{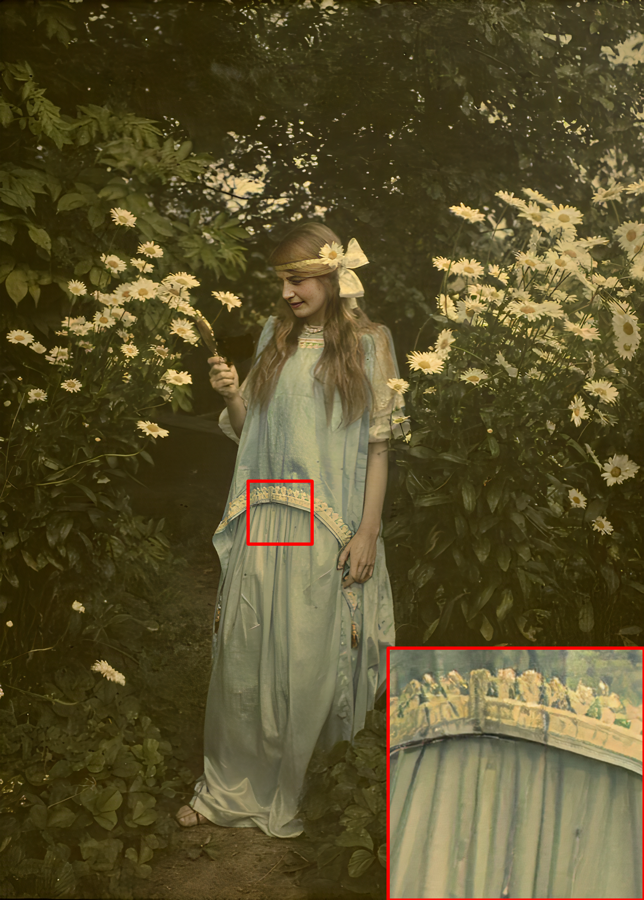}
	\end{minipage}

    \begin{minipage}{0.13\linewidth}
        \centering
        Real-world LR
    \end{minipage}
    \begin{minipage}{0.13\linewidth}
        \centering
        Real-ESRGAN~\cite{Real-ESRGAN}
    \end{minipage}
    \begin{minipage}{0.13\linewidth}
        \centering
        BSRGAN~\cite{BSRGAN}
    \end{minipage}
    \begin{minipage}{0.13\linewidth}
        \centering
        MM-RealSR~\cite{MM-RealSR}
    \end{minipage}
    \begin{minipage}{0.13\linewidth}
        \centering
        ReDegNet~\cite{ReDegNet}
    \end{minipage}
    \begin{minipage}{0.13\linewidth}
        \centering
        ReDegNet$^\dagger$~\cite{ReDegNet}
    \end{minipage}
    \begin{minipage}{0.13\linewidth}
        \centering
        Ours
    \end{minipage}
    
     \caption{Visual comparison against state-of-the-art blind SR methods on real-world low-quality images in our RF200 dataset. ReDegNet$^\dagger$ means that model is fine-tuned for each image following the official configurations of ReDegNet~\cite{ReDegNet}.}% Note that the results of Ours are produced by MetaF2N with one-step fine-tuning. Please zoom in for better observation.}
    \label{fig:metaf2n_results_realworld}
    \vspace{-1.4em}
\end{figure*}

\subsection{Dataset and Training Details}

\noindent\textbf{Training Data.}
For synthesizing data to train the proposed MetaF2N, high-quality facial and natural images are required for inner and outer loop, respectively.
It is worth noting that, the requirement on the training data is that $\mathbf{I}^{\scriptsize\smiley{}}_\mathit{LR}$ and $\mathbf{I}_\mathit{LR}$ share the identical degradation, but they are not necessarily from the same image.
Therefore, to better leverage existing high-quality facial and natural image datasets, we use the first 30,000 images of the aligned {FFHQ}~\cite{FFHQ} dataset for the inner loop, and adopt {DF2K}\footnote{{DF2K} is the combination of {DIV2K}~\cite{DIV2K} and {Flickr2K}~\cite{Flickr2K} datasets.} with 3,450 images in total for the outer loop following recent blind image SR methods~\cite{Real-ESRGAN,BSRGAN,ReDegNet}.
For a fair comparison against other methods, we follow the degradation setting of Real-ESRGAN~\cite{Real-ESRGAN} for low-quality image synthesis.
More details are given in the supplementary material.

\vspace{0.5em}
\noindent\textbf{Testing Data.}
To comprehensively evaluate the proposed MetaF2N, we have constructed several testing datasets.

\emph{Synthetic datasets} are built upon the in-the-wild version of FFHQ~\cite{FFHQ} and CelebA~\cite{CelebA}, where 1,000 high-quality images are extracted from each dataset, and faces occupy around 10\% areas of the whole image on average.
With the high-quality images, we first construct two test datasets whose degradations are independent and identically distributed as the training set, which are denoted by FFHQ$_\mathit{iid}$ and CelebA$_\mathit{iid}$, respectively.
For evaluating the generalization ability on out-of-distribution degradations, we further construct FFHQ$_\mathit{ood}$ and CelebA$_\mathit{ood}$ by changing the parameters of the degradation model, \eg, Gaussian blur $\rightarrow$ motion blur, Gaussian/Poisson noise $\rightarrow$ Speckle noise. 
The detailed parameters are in the supplementary material.

A \emph{real-world dataset} RealFaces200 (RF200) is also established for evaluation under a real scenario, which contains 200 real-world low-quality images collected from the Internet or existing datasets (\eg, WIDER FACE~\cite{WIDERFACE}), and there are one or multiple faces in each image.

\begin{table*}[h]
    \centering
    \small
    \setlength\abovecaptionskip{0.25\baselineskip} 
    \setlength\belowcaptionskip{0.3\baselineskip}
    \setlength\tabcolsep{1mm}
    %\resizebox{\textwidth}{!}{
    \caption{Quantitative comparison on synthetic datasets with independent and identically distributed degradations as the training set. We also show the datasets used for training each method. Note that RealFaces used by ReDegNet~\cite{ReDegNet} consists of 10,000 real-world low-quality face images collected by the authors. The numbers in the parentheses are the image-specific fine-tuning steps of our MetaF2N during inference. The best and second-best results are highlighted by {\color{red}\textbf{bold}} and {\color{blue}\underline{underline}}.}
    
    \label{table:degradtion1}
    \scalebox{.95}{
        \begin{tabular}{cccccccccc}
        \toprule
    
        \multirow{2}{*}{Methods} & \multirow{2}{*}{Training Set} & \multicolumn{4}{c}{FFHQ$_\mathit{iid}$} & \multicolumn{4}{c}{CelebA$_\mathit{iid}$}\\ 
        \cmidrule(r){3-6} \cmidrule(l){7-10} 
         & & PSNR$\uparrow$ & LPIPS$\downarrow$ & FID$\downarrow$ & NIQE$\downarrow$ & PSNR$\uparrow$ & LPIPS$\downarrow$ & FID$\downarrow$ & NIQE$\downarrow$\\ %& PSNR↑ & LPIPS↓ & FID↓ & NIQE↓ & PSNR↑ & LPIPS↓ & FID↓ & NIQE↓ & KID↓ & NIQE↓ & KID↓ & NIQE↓ \\
         \midrule
    
        ESRGAN~\cite{ESRGAN} & {DF2K~\cite{DIV2K, Flickr2K}}, {OST~\cite{OST}} & 25.10 & 0.642 & 123.67 & 7.56 & 24.78 & 0.555 & 99.10 & 6.27 \\
        RealSR~\cite{RealSR-Kernel} & {DF2K~\cite{DIV2K, Flickr2K}} & 25.39 & 0.597 & 122.23 & 7.59 & 24.74 & 0.549 & 99.46 & 5.77 \\
        % DRealSR~\cite{DRealSR} & {DRealSR} & 22.70 & 0.638 & 118.10 & 9.13 & 21.95 & 0.658 & 116.79 & 9.11 \\
        Real-ESRGAN~\cite{Real-ESRGAN} & {DF2K~\cite{DIV2K, Flickr2K}}, {OST~\cite{OST}} & \textcolor{red}{\textbf{26.35}} & 0.293 & 50.70 & 4.85 & \textcolor{blue}{\underline{25.81}} & 0.311 & 51.69 & 5.00 \\
        BSRGAN~\cite{BSRGAN} & {DF2K~\cite{DIV2K, Flickr2K}}, {FFHQ~\cite{FFHQ}}, {WED~\cite{WED}} & 26.27 & 0.305 & 50.98 & 4.63 & \textcolor{red}{\textbf{25.86}} & \textcolor{blue}{\underline{0.309}} & 50.05 & 4.83 \\
        MM-RealSR~\cite{MM-RealSR} & {DF2K~\cite{DIV2K, Flickr2K}}, {OST~\cite{OST}}  & 25.39 & 0.295 & 50.83 & 4.58 & 24.94 & 0.311 & 52.30 & 4.76 \\
        ReDegNet~\cite{ReDegNet} & {DF2K~\cite{DIV2K, Flickr2K}}, {FFHQ~\cite{FFHQ}}, {RealFaces} & 25.64 & 0.309 & 57.04 & 4.79 & 25.17 & 0.329 & 56.84 & 4.94 \\
       
        Ours (1) & {DF2K~\cite{DIV2K, Flickr2K}}, {FFHQ~\cite{FFHQ}} & 26.13 & 0.281 & 45.22 & \textcolor{red}{\textbf{3.81}} & 25.63 & \textcolor{red}{\textbf{0.289}} & 45.72 & \textcolor{red}{\textbf{3.97}} \\
        Ours (10) & {DF2K~\cite{DIV2K, Flickr2K}}, {FFHQ~\cite{FFHQ}} & 26.22 & \textcolor{blue}{\underline{0.280}} & \textcolor{blue}{\underline{44.94}} & \textcolor{blue}{\underline{3.87}} & 25.70 & \textcolor{red}{\textbf{0.289}} & \textcolor{blue}{\underline{45.43}} & \textcolor{blue}{\underline{4.03}} \\
        Ours (20) & {DF2K~\cite{DIV2K, Flickr2K}}, {FFHQ~\cite{FFHQ}} & \textcolor{blue}{\underline{26.30}} & \textcolor{red}{\textbf{0.279}} & \textcolor{red}{\textbf{44.51}} & 3.94 & 25.76 & \textcolor{red}{\textbf{0.289}} & \textcolor{red}{\textbf{45.22}} & 4.10 \\
         \bottomrule
        \end{tabular}
        }
        \vspace{-0.4em}
\end{table*}

\begin{table*}[h]
    \centering
    \small
    \setlength\abovecaptionskip{0.25\baselineskip} 
    \setlength\belowcaptionskip{0.3\baselineskip}
    \setlength\tabcolsep{1mm}
    %\resizebox{\textwidth}{!}{
    \caption{Quantitative comparison on synthetic datasets with out-of-distribution degradations and the collected real-world dataset RF200. ReDegNet$^\dagger$ means that model is fine-tuned for each image following the official configurations of ReDegNet~\cite{ReDegNet}. The numbers in the parentheses are the image-specific fine-tuning steps of our MetaF2N during inference. The best and second-best results are highlighted by {\color{red}\textbf{bold}} and {\color{blue}\underline{underline}}.}
    
    \label{table:degradation2}
    \scalebox{.95}{
        \begin{tabular}{ccccccccccc}
        \toprule
      
        \multirow{2}{*}{Methods} & \multicolumn{4}{c}{FFHQ$_\mathit{ood}$} & \multicolumn{4}{c}{CelebA$_\mathit{ood}$} & \multicolumn{2}{c}{RF200}  \\
        \cmidrule(r){2-5} \cmidrule(l){6-9} \cmidrule(lr){10-11} 
         & PSNR$\uparrow$ & LPIPS$\downarrow$ & FID$\downarrow$ & NIQE$\downarrow$ & PSNR$\uparrow$ & LPIPS$\downarrow$ & FID$\downarrow$ & NIQE$\downarrow$ &  KID$\downarrow$ & NIQE$\downarrow$  \\
         \midrule

        ESRGAN~\cite{ESRGAN} & 24.69 & 0.659 & 101.43 & 6.94 & 24.07 & 0.567 & 87.61 & 5.64 & 21.8 & 5.32  \\
        RealSR~\cite{RealSR-Kernel} & 24.91 & 0.587 & 97.42 & 6.33 & 23.98 & 0.567 & 90.30 & 5.51 & 21.9 & 5.00  \\
  
        Real-ESRGAN~\cite{Real-ESRGAN} & \textcolor{red}{\textbf{25.73}} & 0.302 & 47.87 & 5.34 & \textcolor{blue}{\underline{25.07}} & 0.322 & 48.77 & 5.43 & 22.4 & 3.82  \\
        BSRGAN~\cite{BSRGAN} & \textcolor{red}{\textbf{25.73}} & 0.298 & 46.40 & 4.78 & \textcolor{red}{\textbf{25.33}} & \textcolor{blue}{\underline{0.313}} & 48.12 & 4.91 & 22.1 & 4.11  \\
        MM-RealSR~\cite{MM-RealSR} & 25.03 & 0.297 & 47.30 & 4.99 & 24.41 & 0.317 & 48.76 & 5.10 & 22.1 & 4.12   \\
        ReDegNet~\cite{ReDegNet} & 25.54 & 0.304 & 48.47 & 5.09 & 24.76 & 0.331 & 52.33 & 5.28 & 23.5 & 4.07  \\
        ReDegNet$^\dagger$~\cite{ReDegNet} & - & - & - & - & - & - & - & - & 23.8 & 3.56  \\
      
       Ours (1) & 25.49 & \textcolor{blue}{\underline{0.284}} & 45.07 & \textcolor{red}{\textbf{4.22}} & 24.87 & \textcolor{red}{\textbf{0.297}} & 47.03 & \textcolor{red}{\textbf{4.32}} & 21.2 & \textcolor{red}{\textbf{3.03}}  \\
        Ours (10) & 25.57  & \textcolor{blue}{\underline{0.284}} & \textcolor{blue}{\underline{44.64}} & \textcolor{blue}{\underline{4.30}} & 24.93 & \textcolor{red}{\textbf{0.297}} & \textcolor{blue}{\underline{46.50}} & \textcolor{blue}{\underline{4.39}} & \textcolor{blue}{\underline{21.0}} & \textcolor{blue}{\underline{3.08}} \\
       Ours (20) & \textcolor{blue}{\underline{25.65}} & \textcolor{red}{\textbf{0.283}} & \textcolor{red}{\textbf{44.30}} & 4.36 & 25.00 & \textcolor{red}{\textbf{0.297}} & \textcolor{red}{\textbf{46.23}} & 4.47 & \textcolor{red}{\textbf{20.8}} & 3.12 \\
         \bottomrule
        \end{tabular}
    }
    \vspace{-0.4em}
\end{table*}

\vspace{0.5em}
\noindent\textbf{Implementation Details.}
The proposed MetaF2N mainly focuses on leveraging meta-learning for efficient model adaptation for blind image SR, which is independent of network architecture and can be incorporated into arbitrary second-order differentiable models.
Therefore, we follow Real-ESRGAN~\cite{Real-ESRGAN}, BSRGAN~\cite{BSRGAN}, and ReDegNet~\cite{ReDegNet} to adopt the architecture of ESRGAN~\cite{ESRGAN}.
For training, each image is cropped into patches, and the patch size is 128$\times$128 for the inner loop (\ie, $\mathbf{I}^{\iftoggle{useemoji}{\scriptsize\smiley{}}{\mathit{face}}}_\mathit{LR}$, $\mathbf{I}^{\iftoggle{useemoji}{\scriptsize\smiley{}}{\mathit{face}}}_\mathit{BFR}$, and $\mathbf{I}^{\iftoggle{useemoji}{\scriptsize\smiley{}}{\mathit{face}}}_\mathit{SR}$) and 256$\times$256 for the outer loop (\ie, $\mathbf{I}_\mathit{LR}$, $\mathbf{I}_\mathit{SR}$, and $\mathbf{I}$).
The Adam~\cite{Adam} optimizer with $\beta_1$ = 0.5 and $\beta_2$ = 0.999 is adopted.
For the SR model $\mathit{f}$, the learning rate (lr) of the inner and outer loops are $1\times10^{-2}$ and $3\times10^{-5}$, respectively, while the lr for the  discriminator and MaskNet are $1\times10^{-4}$.
\if 0
The experiments are conducted with TensorFlow~\cite{TensorFlow} on a server with Nvidia RTX A6000 GPUs.
\fi
{The MetaF2N is trained for one week and all experiments are conducted on a server with one RTX A6000 GPU.}

\subsection{Comparison with State-of-the-art Methods}
To show the effectiveness of the proposed MetaF2N, we compare with several state-of-the-art blind image SR methods, including ESRGAN~\cite{ESRGAN}, RealSR~\cite{RealSR}, Real-ESRGAN~\cite{Real-ESRGAN}, BSRGAN~\cite{BSRGAN}, MM-RealSR~\cite{MM-RealSR}, and ReDegNet~\cite{ReDegNet}.
For quantitative evaluation, we utilize PSNR, LPIPS~\cite{LPIPS}, FID~\cite{FID}, and NIQE~\cite{NIQE} for synthetic datasets.
As for the real-world dataset, since no ground-truth is available, PSNR and LPIPS are omitted.
It is worth noting that KID is more accurate and appropriate than FID when evaluating the quality of images with a smaller amount of samples~\cite{KID}. Thus we use KID~\cite{KID} and NIQE~\cite{NIQE} for quantitative evaluation on our RF200 dataset.
To accurately calculate the distance between SR results and real-world high-quality images (\ie, FID and KID), we construct these types of reference images following~\cite{BlurDetection}, which estimates a blurriness score based on the total variance of the Laplacian of an image.
Finally, 3,808 high-quality images are selected from the in-the-wild version of FFHQ~\cite{FFHQ}, whose scores are higher than the threshold defined in~\cite{BlurDetection}.

\vspace{0.5em}
\noindent\textbf{Quantitative Comparison.}
The quantitative evaluation results are provided in \cref{table:degradtion1,table:degradation2}.
We provide three results of our Meta-F2N, \ie, Ours (1), Ours (10), and Ours (20), where the number in the parentheses denotes the amount of fine-tuning steps during inference.
All three models are initialized with the same parameter $\theta$.
For the degradations independent and identically distributed as the training set (\ie, FFHQ$_\mathit{iid}$ and CelebA$_\mathit{iid}$), our MetaF2N can achieve superior performance on LPIPS, FID, and NIQE with only one fine-tuning step, and can surpass most of the competing methods on PSNR.
Besides, with more fine-tuning iterations, the image quality can be further enhanced.
For out-of-distribution degradations in \cref{table:degradation2} (\ie, FFHQ$_\mathit{ood}$, CelebA$_\mathit{ood}$, and RF200), the trends are similar to \cref{table:degradtion1}, and our MetaF2N again outperforms others on most metrics, especially on the real-world RF200 dataset.

As for the competing methods, ESRGAN~\cite{ESRGAN} and RealSR~\cite{RealSR} are trained with simple degradation models, which lead to unsatisfactory performance under more complex and realistic degradations.
Real-ESRGAN~\cite{Real-ESRGAN}, BSRGAN~\cite{BSRGAN}, MM-RealSR~\cite{MM-RealSR}, and ReDegNet~\cite{ReDegNet} leverage more realistic degradation models or even real-world degradations, which contribute to much better results, yet their performance is still limited by the deterministic model and fixed degradation range.
Furthermore, we also fine-tune ReDegNet~\cite{ReDegNet} on RF200, and the results are provided in \cref{table:degradation2} (ReDegNet$^\dagger$).
Compared to MetaF2N that achieves decent performance with only one fine-tuning step, marginal improvements are observed on NIQE even the ReDegNet~\cite{ReDegNet} is carefully fine-tuned hundreds iterations for each image.

\vspace{0.5em}
\noindent\textbf{Qualitative Comparison.}
Apart from the quantitative comparison, we also compare our MetaF2N with state-of-the-art blind image SR methods qualitatively.
Due to the space limit, we show the results generated by Ours (1), which is fine-tuned for 1 step with the face regions, and more qualitative results (including those generated by Ours (10) and Ours (20)) are provided in the supplementary material.
As shown in \cref{fig:metaf2n_results_synthetic,fig:metaf2n_results_realworld}, our results are much clearer and contain more photo-realistic textures, which can be ascribed to the effectiveness of our image-specific fine-tuning scheme.

\subsection{The Visual Results of MaskNet}
\begin{figure}
	\centering
        \footnotesize
        \setlength\abovecaptionskip{0.2\baselineskip}
        \setlength\belowcaptionskip{0.2\baselineskip}
        \begin{minipage}{0.18\linewidth}
		\centering
		\includegraphics[width=1.0\linewidth]{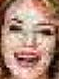}
	\end{minipage}
 	\begin{minipage}{0.18\linewidth}
		\centering
		\includegraphics[width=1.0\linewidth]{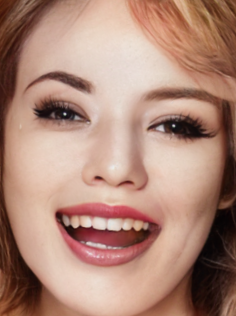}
	\end{minipage}
 	\begin{minipage}{0.18\linewidth}
		\centering
		\includegraphics[width=1.0\linewidth]{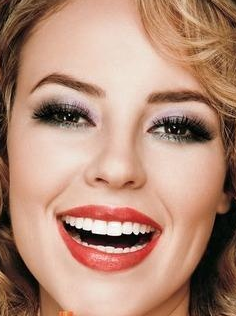}
	\end{minipage}
	\begin{minipage}{0.18\linewidth}
		\centering
		\includegraphics[width=1.0\linewidth]{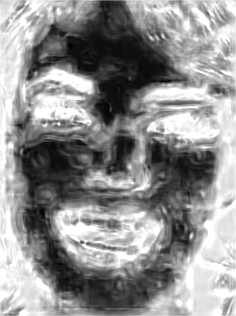}
	\end{minipage}
 	\begin{minipage}{0.18\linewidth}
		\centering
		\includegraphics[width=1.0\linewidth]{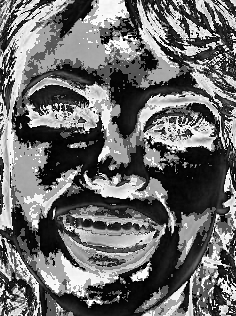}
	\end{minipage}

                \begin{minipage}{0.18\linewidth}
		\centering
		\includegraphics[width=1.0\linewidth]{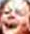}
	\end{minipage}
 	\begin{minipage}{0.18\linewidth}
		\centering
		\includegraphics[width=1.0\linewidth]{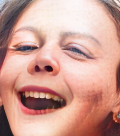}
	\end{minipage}
 	\begin{minipage}{0.18\linewidth}
		\centering
		\includegraphics[width=1.0\linewidth]{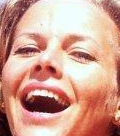}
	\end{minipage}
	\begin{minipage}{0.18\linewidth}
		\centering
		\includegraphics[width=1.0\linewidth]{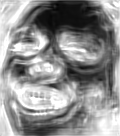}
	\end{minipage}
 	\begin{minipage}{0.18\linewidth}
		\centering
		\includegraphics[width=1.0\linewidth]{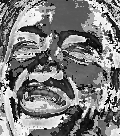}
	\end{minipage}

         \begin{minipage}{0.18\linewidth}
		\centering
		\includegraphics[width=1.0\linewidth]{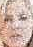}
	\end{minipage}
 	\begin{minipage}{0.18\linewidth}
		\centering
		\includegraphics[width=1.0\linewidth]{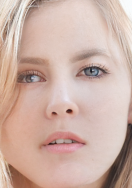}
	\end{minipage}
 	\begin{minipage}{0.18\linewidth}
		\centering
		\includegraphics[width=1.0\linewidth]{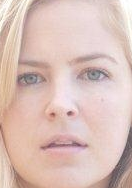}
	\end{minipage}
	\begin{minipage}{0.18\linewidth}
		\centering
		\includegraphics[width=1.0\linewidth]{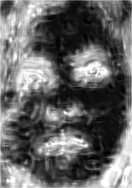}
	\end{minipage}
 	\begin{minipage}{0.18\linewidth}
		\centering
		\includegraphics[width=1.0\linewidth]{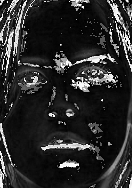}
	\end{minipage}

         \begin{minipage}{0.18\linewidth}
		\centering
		\includegraphics[width=1.0\linewidth]{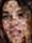}
	\end{minipage}
 	\begin{minipage}{0.18\linewidth}
		\centering
		\includegraphics[width=1.0\linewidth]{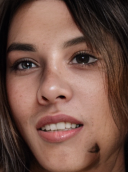}
	\end{minipage}
 	\begin{minipage}{0.18\linewidth}
		\centering
		\includegraphics[width=1.0\linewidth]{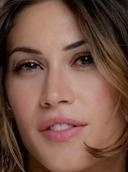}
	\end{minipage}
	\begin{minipage}{0.18\linewidth}
		\centering
		\includegraphics[width=1.0\linewidth]{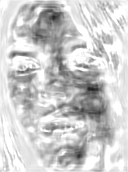}
	\end{minipage}
 	\begin{minipage}{0.18\linewidth}
		\centering
		\includegraphics[width=1.0\linewidth]{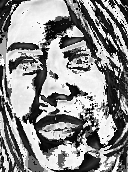}
	\end{minipage}

         \begin{minipage}{0.18\linewidth}
		\centering
		\includegraphics[width=1.0\linewidth]{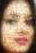}
	\end{minipage}
 	\begin{minipage}{0.18\linewidth}
		\centering
		\includegraphics[width=1.0\linewidth]{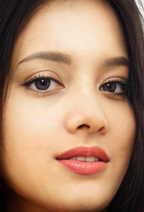}
	\end{minipage}
 	\begin{minipage}{0.18\linewidth}
		\centering
		\includegraphics[width=1.0\linewidth]{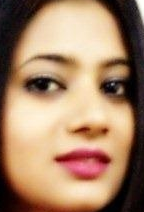}
	\end{minipage}
	\begin{minipage}{0.18\linewidth}
		\centering
		\includegraphics[width=1.0\linewidth]{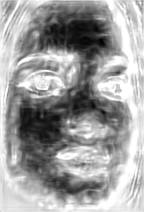}
	\end{minipage}
 	\begin{minipage}{0.18\linewidth}
		\centering
		\includegraphics[width=1.0\linewidth]{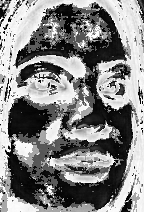}
	\end{minipage}

         \begin{minipage}{0.18\linewidth}
		\centering
		\includegraphics[width=1.0\linewidth]{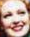}
	\end{minipage}
 	\begin{minipage}{0.18\linewidth}
		\centering
		\includegraphics[width=1.0\linewidth]{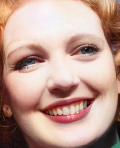}
	\end{minipage}
 	\begin{minipage}{0.18\linewidth}
		\centering
		\includegraphics[width=1.0\linewidth]{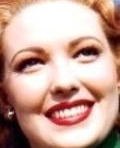}
	\end{minipage}
	\begin{minipage}{0.18\linewidth}
		\centering
		\includegraphics[width=1.0\linewidth]{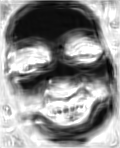}
	\end{minipage}
 	\begin{minipage}{0.18\linewidth}
		\centering
		\includegraphics[width=1.0\linewidth]{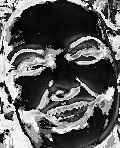}
	\end{minipage}

    \begin{minipage}{0.18\linewidth}
        \centering
        LR
    \end{minipage}
    \begin{minipage}{0.18\linewidth}
        \centering
        GPEN~\cite{GPEN}
    \end{minipage}
    \begin{minipage}{0.18\linewidth}
        \centering
        GT
    \end{minipage}
    \begin{minipage}{0.18\linewidth}
        \centering
        $\mathbf{m}$ 
    \end{minipage}
    \begin{minipage}{0.18\linewidth}
        \centering
        $1-\mathit{EM}$
    \end{minipage}
    
     \caption{Visual results of MaskNet.}
    \label{fig:mask_vis}
    \vspace{-2em}
\end{figure}

To better explain the effect of our MaskNet, we calculate the error map $\mathit{EM}$ between ground-truth faces $\mathbf{I}^{\iftoggle{useemoji}{\scriptsize\smiley{}}{\mathit{face}}}$ and GPEN~\cite{GPEN} generated faces $\mathbf{I}^{\iftoggle{useemoji}{\scriptsize\smiley{}}{\mathit{face}}}_\mathit{BFR}$ directly through subtraction and normalization, \ie,
\begin{equation}
    \mathit{EM} =\frac{|\mathbf{I}^{\iftoggle{useemoji}{\scriptsize\smiley{}}{\mathit{face}}} - \mathbf{I}^{\iftoggle{useemoji}{\scriptsize\smiley{}}{\mathit{face}}}_\mathit{BFR}|}{\max(|\mathbf{I}^{\iftoggle{useemoji}{\scriptsize\smiley{}}{\mathit{face}}} - \mathbf{I}^{\iftoggle{useemoji}{\scriptsize\smiley{}}{\mathit{face}}}_\mathit{BFR}|)}.
\end{equation}
In \cref{fig:mask_vis}, we show the predicted $\mathbf{m}$ and $1-\mathit{EM}$.
One can see that the predicted $\mathbf{m}$ follows similar distribution to $1-\mathit{EM}$, indicating that our MaskNet has the ability to predict the gap between generated faces and real ones, which benefits the blind image SR performance.
Besides, the predicted $\mathbf{m}$ is smoother than $1-\mathit{EM}$, due to that
1)~$\mathbf{m}$ is predicted from ($\mathbf{I}^{\scriptsize\smiley{}}_\mathit{LR}$ and $\mathbf{I}^{\scriptsize\smiley{}}_\mathit{BFR}$), which lead to less accurate results, and
2)~a regularization term $\mathcal{L}_\mathrm{reg}=\|\mathbf{m}-1\|_2$ is applied to constrain $\mathbf{m}$ not to deviate too far from 1.

\subsection{Ablation Studies}
To verify the effectiveness of the components of the proposed MetaF2N, we have also conducted several ablation studies.
Without loss of generality, all ablation studies are performed with FFHQ$_\mathit{iid}$ based on Ours (1).
Due to the space limit, qualitative results of ablation studies are given in the supplementary material.

\vspace{0.5em}
\noindent\textbf{MaskNet Configuration.}
To show the effectiveness of the MaskNet, we train a variant of our MetaF2N by removing the MaskNet.
Besides, as introduced in \cref{sec:metaf2n_introduction}, we design the MaskNet inspired by degraded-reference IQA methods, which takes both low-quality and restored faces as input (denoted by MaskNet$_\mathit{DR}$).
For verifying the design, we also consider another variant of MetaF2N which utilizes the non-reference scheme, \ie, only the restored face is taken by the MaskNet (denoted by MaskNet$_\mathit{NR}$).
As shown in \cref{table:masknet}, the deployment of MaskNet brings considerable improvements, and utilizing the low-quality face image as a reference is also useful for our task.

\vspace{0.5em}
\noindent\textbf{Training Data Configuration.}
Since ground-truth faces are available during training, we train another variant by using the ground-truth face image as the inner loop supervision.
Note that the MaskNet is trained together with the SR model, and we omitted it when training the version using ground-truth faces as inner loop supervision since the optimal solution should be an all-one mask.
As shown in \cref{table:masknet}, the performance decreases by a large margin compared to MetaF2N, which is attributed to the gaps in the inner loop supervision between the training and inference phases.

\vspace{0.5em}
\noindent\textbf{Inner Loop Supervision Scheme.}
Since there are often multiple faces in a single image, as a natural extension of our method, it is intuitive to apply multiple faces to stabilize the fine-tuning process.
Therefore, we also explore the inner loop supervision scheme based on another 1,000 images sampled from the in-the-wild version of FFHQ~\cite{FFHQ} with multiple faces.
As shown in \cref{table:multiface}, using patches of multiple faces can slightly benefit the final performance.

\vspace{0.5em}
\noindent\textbf{Influence of Different Face Hallucination Network.}
To find the influence of different face hallucination network for the MetaF2N, we conduct an ablation study by replacing GPEN~\cite{GPEN} with CodeFormer~\cite{CodeFormer} during training and testing. As shown in \cref{tab:face_network}, one can see that our MetaF2N shows satisfactory robustness with different face hallucination networks for both FFHQ$_\mathit{iid}$ and FFHQ$_\mathit{ood}$. Although the CodeFormer~\cite{CodeFormer} can restore higher quaility faces compared with GPEN~\cite{GPEN}, our MaskNet minish the gap through allocating weights for each pixel of restored faces, which enhances the robustness of MetaF2N for different face hallucination networks.

\begin{table}[th]
    \centering
    \small
    \setlength\abovecaptionskip{0.2\baselineskip} 
    \setlength\belowcaptionskip{0.3\baselineskip}
    \setlength\tabcolsep{1mm}
    \renewcommand\arraystretch{1.2}
    \caption{Ablation study on the training scheme regarding data configuration and the MaskNet. Train/Test Faces denote the setting of inner-loop supervision during the training and inference stages. 
    }
  
    \label{table:masknet}
    \scalebox{0.88}{
        \begin{tabular}{ccccccc}
        \toprule
       
        {MaskNet} & {Train Faces} & {Test Faces} & PSNR$\uparrow$ & LPIPS$\downarrow$ & FID$\downarrow$ & NIQE$\downarrow$ \\
         \midrule
        - & GT & GPEN   & 25.68 & 0.285 &  \textcolor{blue}{\underline{46.96}} & \textcolor{blue}{\underline{4.08}}\\
        - & GPEN & GPEN & \textcolor{red}{\textbf{26.17}} &  \textcolor{blue}{\underline{0.284}} & 47.89 & 4.21\\
        MaskNet$_\mathit{NR}$ &  GPEN & GPEN  & 26.07 & \textcolor{blue}{\underline{0.284}} & 47.97 & 4.09 \\
        MaskNet$_\mathit{DR}$ & GPEN & GPEN & \textcolor{blue}{\underline{26.13}} & \textcolor{red}{\textbf{0.281}} & \textcolor{red}{\textbf{45.22}} & \textcolor{red}{\textbf{3.81}} \\
         \bottomrule
    \end{tabular}}
    \vspace{-1em}
\end{table}

\begin{table}
    \centering
    \small
    \setlength\abovecaptionskip{0.2\baselineskip} 
    \setlength\belowcaptionskip{0.3\baselineskip}
    \setlength\tabcolsep{1mm}
    \renewcommand\arraystretch{1.2}
   
    \caption{Ablation study on inner loop supervision, which is performed on 1,000 images with multiple faces ($\sim$2.3 faces/image) from the in-the-wild version of FFHQ~\cite{FFHQ}.
    }
    
    \label{table:multiface}
    \scalebox{0.92}{
        \begin{tabular}{cccccc}
        \toprule
      
        {\# of Faces} & {Data Form} & PSNR$\uparrow$ & LPIPS$\downarrow$ & FID$\downarrow$ & NIQE$\downarrow$  \\

         \midrule

        Single & Full image & 25.73 & \textcolor{blue}{\underline{0.288}} & 42.96 & \textcolor{red}{\textbf{3.63}} \\
        Single & 32 patches & \textcolor{blue}{\underline{25.77}} & \textcolor{blue}{\underline{0.288}} & 42.89 & \textcolor{blue}{\underline{3.65}} \\
        Multiple & 32 patches in total & \textcolor{red}{\textbf{25.79}} & \textcolor{red}{\textbf{0.287}} & \textcolor{blue}{\underline{42.85}} & \textcolor{blue}{\underline{3.65}} \\
        Multiple & 32 patches per face & \textcolor{red}{\textbf{25.79}} & \textcolor{red}{\textbf{0.287}} &\textcolor{red}{\textbf{42.81}} & \textcolor{blue}{\underline{3.65}} \\
         \bottomrule
        \end{tabular}
    }
    \vspace{-1em}
\end{table}
\begin{table}[h]
    \centering
    \small
    \setlength\abovecaptionskip{0.2\baselineskip} 
    \setlength\belowcaptionskip{0.3\baselineskip}
    \setlength\tabcolsep{1mm}
    \renewcommand\arraystretch{1.2}
    % \vspace{-3mm}
    \caption{Ablation study on face hallucination networks， which is performed on both FFHQ$_\mathit{iid}$ and FFHQ$_\mathit{ood}$.
    }
    % \vspace{-3mm}
    \label{tab:face_network}
    \scalebox{0.88}{
        \begin{tabular}{ccccccc}
        \toprule
       
        {Testsets} & {Train Faces} & {Test Faces} & PSNR$\uparrow$ & LPIPS$\downarrow$ & FID$\downarrow$ & NIQE$\downarrow$ \\
         \midrule
        \multirow{2}{*}{FFHQ$_\mathit{iid}$} & GPEN & GPEN   & \textcolor{red}{\textbf{26.13}} & 0.281 & \textcolor{red}{\textbf{45.22}} & \textcolor{red}{\textbf{3.81}} \\
                                             & CodeFormer & CodeFormer & 26.10 & \textcolor{red}{\textbf{0.278}} & 45.83 & 3.83 \\
         \midrule
    
        \multirow{2}{*}{FFHQ$_\mathit{ood}$} & GPEN & GPEN  & 25.49 & 0.284 & 45.07 & \textcolor{red}{\textbf{4.22}} \\
                                             & CodeFormer & CodeFormer & \textcolor{red}{\textbf{25.51}} & \textcolor{red}{\textbf{0.281}} & \textcolor{red}{\textbf{44.58}} & 4.24 \\
         \bottomrule
    \end{tabular}}
    \vspace{-1mm}
\end{table}

\subsection{Limitations and Future Work}
Since our MetaF2N is trained in a MAML framework, the performance before fine-tuning is unconstrained.
As such, our model requires fine-tuning with the contained faces for one or multiple steps, making it cannot be directly applied if no faces exist in real-world low-quality images, which limits the application scenarios of our method.
In the future, we plan to solve this problem by collecting a real-world low-quality training set and providing a better initialization of parameters that can be used to achieve satisfactory performance even without fine-tuning.

\section{Conclusion}
In this paper, we proposed a MetaF2N framework for face-guided blind image SR.
The MetaF2N improves the practicability of leveraging blind face restoration for processing image-specific degradations.
A MaskNet is deployed to predict the loss weights for different positions considering the gaps between recovered faces and potential ground-truth ones.
The proposed MetaF2N achieves superior performance on synthetic datasets with both independent and identically distributed and out-of-distribution degradations compared to training sets.
The effectiveness is also validated on the collected real-world dataset.

\section*{Acknowledgement}
This work was supported in part by the National Key Research and Development Program of China under Grant No. 2022YFA1004103 and the National Natural Science Foundation of China (NSFC) under Grant No. U19A2073.

{\small
\bibliographystyle{ieee_fullname}
\bibliography{egpaper_final}

\begin{thebibliography}{10}\itemsep=-1pt

\bibitem{DIV2K}
Eirikur Agustsson and Radu Timofte.
\newblock {NTIRE} 2017 challenge on single image super-resolution: Dataset and
  study.
\newblock In {\em IEEE Conference on Computer Vision and Pattern Recognition
  Workshops}, pages 126--135, 2017.

\bibitem{DRIQA_ICCV21}
Shahrukh Athar and Zhou Wang.
\newblock Degraded reference image quality assessment.
\newblock {\em IEEE Transactions on Image Processing}, pages 822--837, 2023.

\bibitem{KID}
Miko{\l}aj Bi{\'n}kowski, Danica~J Sutherland, Michael Arbel, and Arthur
  Gretton.
\newblock Demystifying mmd gans.
\newblock In {\em International Conference on Learning Representations}, pages
  1--36, 2018.

\bibitem{BlurDetection}
Will Brennan.
\newblock Blurdetection2.
\newblock \url{https://github.com/WillBrennan/BlurDetection2}, 2017.

\bibitem{RealSR}
Jianrui Cai, Hui Zeng, Hongwei Yong, Zisheng Cao, and Lei Zhang.
\newblock Toward real-world single image super-resolution: A new benchmark and
  a new model.
\newblock In {\em IEEE International Conference on Computer Vision}, pages
  3086--3095, 2019.

\bibitem{GLEAN}
Kelvin~CK Chan, Xintao Wang, Xiangyu Xu, Jinwei Gu, and Chen~Change Loy.
\newblock {GLEAN}: Generative latent bank for large-factor image
  super-resolution.
\newblock In {\em IEEE Conference on Computer Vision and Pattern Recognition},
  pages 14245--14254, 2021.

\bibitem{chen2021progressive}
Chaofeng Chen, Xiaoming Li, Lingbo Yang, Xianhui Lin, Lei Zhang, and Kwan-Yee~K
  Wong.
\newblock Progressive semantic-aware style transformation for blind face
  restoration.
\newblock In {\em IEEE Conference on Computer Vision and Pattern Recognition},
  pages 11896--11905, 2021.

\bibitem{City100}
Chang Chen, Zhiwei Xiong, Xinmei Tian, Zheng-Jun Zha, and Feng Wu.
\newblock Camera lens super-resolution.
\newblock In {\em IEEE Conference on Computer Vision and Pattern Recognition},
  pages 1652--1660, 2019.

\bibitem{Survey_InfoFuse}
Honggang Chen, Xiaohai He, Linbo Qing, Yuanyuan Wu, Chao Ren, Ray~E Sheriff,
  and Ce Zhu.
\newblock Real-world single image super-resolution: A brief review.
\newblock {\em Information Fusion}, pages 124--145, 2022.

\bibitem{DualSR}
Mohammad Emad, Maurice Peemen, and Henk Corporaal.
\newblock {DualSR}: Zero-shot dual learning for real-world super-resolution.
\newblock In {\em IEEE Winter Conference on Applications of Computer Vision},
  pages 1630--1639, 2021.

\bibitem{MAML}
Chelsea Finn, Pieter Abbeel, and Sergey Levine.
\newblock Model-agnostic meta-learning for fast adaptation of deep networks.
\newblock In {\em International Conference on Machine Learning}, pages
  1126--1135, 2017.

\bibitem{FSSR}
Manuel Fritsche, Shuhang Gu, and Radu Timofte.
\newblock Frequency separation for real-world super-resolution.
\newblock In {\em IEEE International Conference on Computer Vision Workshops},
  pages 3599--3608, 2019.

\bibitem{GAN}
Ian Goodfellow, Jean Pouget-Abadie, Mehdi Mirza, Bing Xu, David Warde-Farley,
  Sherjil Ozair, Aaron Courville, and Yoshua Bengio.
\newblock Generative adversarial networks.
\newblock {\em Communications of the ACM}, pages 139--144, 2020.

\bibitem{grant2018recasting}
Erin Grant, Chelsea Finn, Sergey Levine, Trevor Darrell, and Thomas Griffiths.
\newblock Recasting gradient-based meta-learning as hierarchical bayes.
\newblock {\em arXiv preprint arXiv:1801.08930}, 2018.

\bibitem{IKC}
Jinjin Gu, Hannan Lu, Wangmeng Zuo, and Chao Dong.
\newblock Blind super-resolution with iterative kernel correction.
\newblock In {\em IEEE Conference on Computer Vision and Pattern Recognition},
  pages 1604--1613, 2019.

\bibitem{gu2022vqfr}
Yuchao Gu, Xintao Wang, Liangbin Xie, Chao Dong, Gen Li, Ying Shan, and
  Ming-Ming Cheng.
\newblock {VQFR}: Blind face restoration with vector-quantized dictionary and
  parallel decoder.
\newblock In {\em European Conference on Computer Vision}, pages 126--143,
  2022.

\bibitem{FID}
Martin Heusel, Hubert Ramsauer, Thomas Unterthiner, Bernhard Nessler, and Sepp
  Hochreiter.
\newblock Gans trained by a two time-scale update rule converge to a local nash
  equilibrium.
\newblock {\em Advances in Neural Information Processing Systems}, pages
  6626--6637, 2017.

\bibitem{DAN}
Yan Huang, Shang Li, Liang Wang, Tieniu Tan, et~al.
\newblock Unfolding the alternating optimization for blind super resolution.
\newblock {\em Advances in Neural Information Processing Systems}, pages
  5632--5643, 2020.

\bibitem{RealSR-Kernel}
Xiaozhong Ji, Yun Cao, Ying Tai, Chengjie Wang, Jilin Li, and Feiyue Huang.
\newblock Real-world super-resolution via kernel estimation and noise
  injection.
\newblock In {\em IEEE Conference on Computer Vision and Pattern Recognition
  Workshops}, pages 466--467, 2020.

\bibitem{ImagePairs}
Hamid Reza~Vaezi Joze, Ilya Zharkov, Karlton Powell, Carl Ringler, Luming
  Liang, Andy Roulston, Moshe Lutz, and Vivek Pradeep.
\newblock {ImagePairs}: Realistic super resolution dataset via beam splitter
  camera rig.
\newblock In {\em IEEE Conference on Computer Vision and Pattern Recognition
  Workshops}, pages 518--519, 2020.

\bibitem{FFHQ}
Tero Karras, Samuli Laine, and Timo Aila.
\newblock A style-based generator architecture for generative adversarial
  networks.
\newblock In {\em IEEE Conference on Computer Vision and Pattern Recognition},
  pages 4401--4410, 2019.

\bibitem{StyleGAN2}
Tero Karras, Samuli Laine, Miika Aittala, Janne Hellsten, Jaakko Lehtinen, and
  Timo Aila.
\newblock Analyzing and improving the image quality of stylegan.
\newblock In {\em IEEE Conference on Computer Vision and Pattern Recognition},
  pages 8110--8119, 2020.

\bibitem{Adam}
Diederik~P Kingma and Jimmy Ba.
\newblock Adam: A method for stochastic optimization.
\newblock {\em arXiv preprint arXiv:1412.6980}, 2014.

\bibitem{AlexNet}
Alex Krizhevsky, Ilya Sutskever, and Geoffrey~E Hinton.
\newblock Imagenet classification with deep convolutional neural networks.
\newblock {\em Communications of the ACM}, pages 84--90, 2017.

\bibitem{Meta-KernelGAN}
Royson Lee, Rui Li, Stylianos~I Venieris, Timothy Hospedales, Ferenc
  Husz{\'a}r, and Nicholas~D Lane.
\newblock Meta-learned kernel for blind super-resolution kernel estimation.
\newblock {\em arXiv preprint arXiv:2212.07886}, 2022.

\bibitem{ReDegNet}
Xiaoming Li, Chaofeng Chen, Xianhui Lin, Wangmeng Zuo, and Lei Zhang.
\newblock From face to natural image: Learning real degradation for blind image
  super-resolution.
\newblock In {\em European Conference on Computer Vision}, pages 376--392,
  2022.

\bibitem{DFDNet}
Xiaoming Li, Chaofeng Chen, Shangchen Zhou, Xianhui Lin, Wangmeng Zuo, and Lei
  Zhang.
\newblock Blind face restoration via deep multi-scale component dictionaries.
\newblock In {\em European Conference on Computer Vision}, pages 399--415,
  2020.

\bibitem{ASFFNet}
Xiaoming Li, Wenyu Li, Dongwei Ren, Hongzhi Zhang, Meng Wang, and Wangmeng Zuo.
\newblock Enhanced blind face restoration with multi-exemplar images and
  adaptive spatial feature fusion.
\newblock In {\em IEEE Conference on Computer Vision and Pattern Recognition},
  pages 2706--2715, 2020.

\bibitem{GFRNet}
Xiaoming Li, Ming Liu, Yuting Ye, Wangmeng Zuo, Liang Lin, and Ruigang Yang.
\newblock Learning warped guidance for blind face restoration.
\newblock In {\em European Conference on Computer Vision}, pages 272--289,
  2018.

\bibitem{DMDNet}
Xiaoming Li, Shiguang Zhang, Shangchen Zhou, Lei Zhang, and Wangmeng Zuo.
\newblock Learning dual memory dictionaries for blind face restoration.
\newblock {\em IEEE Transactions on Pattern Analysis and Machine Intelligence},
  pages 1--13, 2022.

\bibitem{MARCONet}
Xiaoming Li, Wangmeng Zuo, and Chen~Change Loy.
\newblock Learning generative structure prior for blind text image
  super-resolution.
\newblock In {\em IEEE Conference on Computer Vision and Pattern Recognition},
  pages 10103--10113, 2023.

\bibitem{DAWSON}
Weixin Liang, Zixuan Liu, and Can Liu.
\newblock Dawson: A domain adaptive few shot generation framework.
\newblock {\em arXiv preprint arXiv:2001.00576}, 2020.

\bibitem{Survey_TRAMI22}
Anran Liu, Yihao Liu, Jinjin Gu, Yu Qiao, and Chao Dong.
\newblock Blind image super-resolution: A survey and beyond.
\newblock {\em IEEE Transactions on Pattern Analysis and Machine Intelligence},
  pages 1--19, 2022.

\bibitem{CelebA}
Ziwei Liu, Ping Luo, Xiaogang Wang, and Xiaoou Tang.
\newblock Deep learning face attributes in the wild.
\newblock In {\em IEEE International Conference on Computer Vision}, pages
  3730--3738, 2015.

\bibitem{WED}
Kede Ma, Zhengfang Duanmu, Qingbo Wu, Zhou Wang, Hongwei Yong, Hongliang Li,
  and Lei Zhang.
\newblock Waterloo exploration database: New challenges for image quality
  assessment models.
\newblock {\em IEEE Transactions on Image Processing}, pages 1004--1016, 2016.

\bibitem{PULSE}
Sachit Menon, Alexandru Damian, Shijia Hu, Nikhil Ravi, and Cynthia Rudin.
\newblock {PULSE}: Self-supervised photo upsampling via latent space
  exploration of generative models.
\newblock In {\em IEEE Conference on Computer Vision and Pattern Recognition},
  pages 2437--2445, 2020.

\bibitem{mishra2017simple}
Nikhil Mishra, Mostafa Rohaninejad, Xi Chen, and Pieter Abbeel.
\newblock A simple neural attentive meta-learner.
\newblock {\em arXiv preprint arXiv:1707.03141}, 2017.

\bibitem{NIQE}
Anish Mittal, Rajiv Soundararajan, and Alan~C Bovik.
\newblock Making a “completely blind” image quality analyzer.
\newblock {\em IEEE Signal Processing Letters}, pages 209--212, 2012.

\bibitem{MM-RealSR}
Chong Mou, Yanze Wu, Xintao Wang, Chao Dong, Jian Zhang, and Ying Shan.
\newblock Metric learning based interactive modulation for real-world
  super-resolution.
\newblock In {\em European Conference on Computer Vision}, pages 723--740,
  2022.

\bibitem{oreshkin2018tadam}
Boris Oreshkin, Pau Rodr{\'\i}guez~L{\'o}pez, and Alexandre Lacoste.
\newblock {TADAM}: Task dependent adaptive metric for improved few-shot
  learning.
\newblock {\em Advances in Neural Information Processing Systems}, pages
  719--729, 2018.

\bibitem{MLSR}
Seobin Park, Jinsu Yoo, Donghyeon Cho, Jiwon Kim, and Tae~Hyun Kim.
\newblock Fast adaptation to super-resolution networks via meta-learning.
\newblock In {\em European Conference on Computer Vision}, pages 754--769,
  2020.

\bibitem{ZSSR}
Assaf Shocher, Nadav Cohen, and Michal Irani.
\newblock “zero-shot” super-resolution using deep internal learning.
\newblock In {\em IEEE Conference on Computer Vision and Pattern Recognition},
  pages 3118--3126, 2018.

\bibitem{MZSR}
Jae~Woong Soh, Sunwoo Cho, and Nam~Ik Cho.
\newblock Meta-transfer learning for zero-shot super-resolution.
\newblock In {\em IEEE Conference on Computer Vision and Pattern Recognition},
  pages 3516--3525, 2020.

\bibitem{sun2019meta}
Qianru Sun, Yaoyao Liu, Tat-Seng Chua, and Bernt Schiele.
\newblock Meta-transfer learning for few-shot learning.
\newblock In {\em IEEE Conference on Computer Vision and Pattern Recognition},
  pages 403--412, 2019.

\bibitem{sung2018learning}
Flood Sung, Yongxin Yang, Li Zhang, Tao Xiang, Philip~HS Torr, and Timothy~M
  Hospedales.
\newblock Learning to compare: Relation network for few-shot learning.
\newblock In {\em IEEE Conference on Computer Vision and Pattern Recognition},
  pages 1199--1208, 2018.

\bibitem{Flickr2K}
Radu Timofte, Eirikur Agustsson, Luc Van~Gool, Ming-Hsuan Yang, and Lei Zhang.
\newblock {NTIRE} 2017 challenge on single image super-resolution: Methods and
  results.
\newblock In {\em IEEE Conference on Computer Vision and Pattern Recognition
  Workshops}, pages 114--125, 2017.

\bibitem{DIP}
Dmitry Ulyanov, Andrea Vedaldi, and Victor Lempitsky.
\newblock Deep image prior.
\newblock In {\em IEEE Conference on Computer Vision and Pattern Recognition},
  pages 9446--9454, 2018.

\bibitem{vinyals2016matching}
Oriol Vinyals, Charles Blundell, Timothy Lillicrap, Daan Wierstra, et~al.
\newblock Matching networks for one shot learning.
\newblock {\em Advances in Neural Information Processing Systems}, pages
  3630--3638, 2016.

\bibitem{DASR}
Longguang Wang, Yingqian Wang, Xiaoyu Dong, Qingyu Xu, Jungang Yang, Wei An,
  and Yulan Guo.
\newblock Unsupervised degradation representation learning for blind
  super-resolution.
\newblock In {\em IEEE Conference on Computer Vision and Pattern Recognition},
  pages 10581--10590, 2021.

\bibitem{GFPGAN}
Xintao Wang, Yu Li, Honglun Zhang, and Ying Shan.
\newblock Towards real-world blind face restoration with generative facial
  prior.
\newblock In {\em IEEE Conference on Computer Vision and Pattern Recognition},
  pages 9168--9178, 2021.

\bibitem{Real-ESRGAN}
Xintao Wang, Liangbin Xie, Chao Dong, and Ying Shan.
\newblock {Real-ESRGAN}: Training real-world blind super-resolution with pure
  synthetic data.
\newblock In {\em IEEE International Conference on Computer Vision}, pages
  1905--1914, 2021.

\bibitem{OST}
Xintao Wang, Ke Yu, Chao Dong, and Chen~Change Loy.
\newblock Recovering realistic texture in image super-resolution by deep
  spatial feature transform.
\newblock In {\em IEEE Conference on Computer Vision and Pattern Recognition},
  pages 606--615, 2018.

\bibitem{ESRGAN}
Xintao Wang, Ke Yu, Shixiang Wu, Jinjin Gu, Yihao Liu, Chao Dong, Yu Qiao, and
  Chen Change~Loy.
\newblock {ESRGAN}: Enhanced super-resolution generative adversarial networks.
\newblock In {\em European Conference on Computer Vision Workshops}, pages
  63--79, 2018.

\bibitem{wang2022restoreformer}
Zhouxia Wang, Jiawei Zhang, Runjian Chen, Wenping Wang, and Ping Luo.
\newblock {RestoreFormer}: High-quality blind face restoration from undegraded
  key-value pairs.
\newblock In {\em IEEE Conference on Computer Vision and Pattern Recognition},
  pages 17491--17500, 2022.

\bibitem{DRealSR}
Pengxu Wei, Ziwei Xie, Hannan Lu, Zongyuan Zhan, Qixiang Ye, Wangmeng Zuo, and
  Liang Lin.
\newblock Component divide-and-conquer for real-world image super-resolution.
\newblock In {\em European Conference on Computer Vision}, pages 101--117,
  2020.

\bibitem{WIDERFACE}
Shuo Yang, Ping Luo, Chen~Change Loy, and Xiaoou Tang.
\newblock Wider face: A face detection benchmark.
\newblock In {\em IEEE Conference on Computer Vision and Pattern Recognition},
  pages 5525--5533, 2016.

\bibitem{GPEN}
Tao Yang, Peiran Ren, Xuansong Xie, and Lei Zhang.
\newblock Gan prior embedded network for blind face restoration in the wild.
\newblock In {\em IEEE Conference on Computer Vision and Pattern Recognition},
  pages 672--681, 2021.

\bibitem{CinCGAN}
Yuan Yuan, Siyuan Liu, Jiawei Zhang, Yongbing Zhang, Chao Dong, and Liang Lin.
\newblock Unsupervised image super-resolution using cycle-in-cycle generative
  adversarial networks.
\newblock In {\em IEEE Conference on Computer Vision and Pattern Recognition
  Workshops}, pages 701--710, 2018.

\bibitem{BSRGAN}
Kai Zhang, Jingyun Liang, Luc Van~Gool, and Radu Timofte.
\newblock Designing a practical degradation model for deep blind image
  super-resolution.
\newblock In {\em IEEE International Conference on Computer Vision}, pages
  4791--4800, 2021.

\bibitem{LPIPS}
Richard Zhang, Phillip Isola, Alexei~A Efros, Eli Shechtman, and Oliver Wang.
\newblock The unreasonable effectiveness of deep features as a perceptual
  metric.
\newblock In {\em IEEE Conference on Computer Vision and Pattern Recognition},
  pages 586--595, 2018.

\bibitem{SR-RAW}
Xuaner Zhang, Qifeng Chen, Ren Ng, and Vladlen Koltun.
\newblock Zoom to learn, learn to zoom.
\newblock In {\em IEEE Conference on Computer Vision and Pattern Recognition},
  pages 3762--3770, 2019.

\bibitem{DRIQA_TIP23}
Heliang Zheng, Huan Yang, Jianlong Fu, Zheng-Jun Zha, and Jiebo Luo.
\newblock Learning conditional knowledge distillation for degraded-reference
  image quality assessment.
\newblock In {\em IEEE International Conference on Computer Vision}, pages
  10242--10251, 2021.

\bibitem{CodeFormer}
Shangchen Zhou, Kelvin~CK Chan, Chongyi Li, and Chen~Change Loy.
\newblock Towards robust blind face restoration with codebook lookup
  transformer.
\newblock {\em arXiv preprint arXiv:2206.11253}, 2022.

\bibitem{zhu2022blind}
Feida Zhu, Junwei Zhu, Wenqing Chu, Xinyi Zhang, Xiaozhong Ji, Chengjie Wang,
  and Ying Tai.
\newblock Blind face restoration via integrating face shape and generative
  priors.
\newblock In {\em IEEE Conference on Computer Vision and Pattern Recognition},
  pages 7662--7671, 2022.

\end{thebibliography}
}

\end{document}